\documentclass[10pt,journal,compsoc]{IEEEtran}
\ifCLASSOPTIONcompsoc
\usepackage[nocompress]{cite}
\else
\usepackage{cite}
\fi

\ifCLASSINFOpdf
\usepackage[pdftex]{graphicx}
\usepackage{multirow}
\usepackage{amsmath}
\usepackage{amssymb}
\usepackage{caption}
\newcommand{\tabincell}[2]{\begin{tabular}{@{}#1@{}}#2\end{tabular}}
\usepackage{xcolor} 
\usepackage{color} 
\usepackage{makecell}
\usepackage{pifont}
\usepackage[colorlinks,linkcolor=red]{hyperref}
\usepackage{verbatim} 
\usepackage{comment}
\usepackage{booktabs}
\usepackage{subfigure}
\usepackage{ragged2e}

\else
\fi

\begin{document}
	%
	\title{Deep Lookup Network}
	
	\author{Yulan Guo, Longguang Wang, Wendong Mao, Xiaoyu Dong, Yingqian Wang, Li Liu, Wei An
		\IEEEcompsocitemizethanks{\IEEEcompsocthanksitem {Yulan Guo is with the School of Electronics and Communication Engineering, Shenzhen Campus of Sun Yat-sen University, Sun Yat-sen University, Shenzhen 518107, China.} E-mail: guoyulan@sysu.edu.cn.
		\IEEEcompsocthanksitem Longguang Wang is with the Aviation University of Air Force, Changchun 130022, China. E-mail: wanglongguang15@nudt.edu.cn.
		\IEEEcompsocthanksitem Wendong Mao is with the College of 
		Integrated Circuits, Shenzhen Campus of Sun Yat-sen University, Sun Yat-sen University, Shenzhen 518107, China. E-mail: maowd@mail.sysu.edu.cn.
		\IEEEcompsocthanksitem Xiaoyu Dong is with the University of Tokyo, Tokyo 113-8654, Japan. E-mail: dong@ms.k.utokyo.ac.jp.
		\IEEEcompsocthanksitem Yingqian Wang, Li Liu, and Wei An are with the College of Electronic Science and Technology, National University of Defense Technology, Changsha 410073, China. E-mail: wangyingqian16@nudt.edu.cn, dreamliu2010@gmail.com, anwei@nudt.edu.cn.
		\IEEEcompsocthanksitem Corresponding author: Longguang Wang. 
		}
	}

	\markboth{IEEE Transactions on Pattern Analysis and Machine Intelligence,~Vol.~XX, No.~XX, XX~2020}%
	{Wang \MakeLowercase{\textit{et al.}}: XXX}

	\IEEEtitleabstractindextext{%
		\begin{abstract}
			\justifying   
			Convolutional neural networks are constructed with massive operations with different types and are highly computationally intensive. Among these operations, multiplication operation is higher in computational complexity and usually requires {more} energy consumption with longer inference time than other operations, which hinders the deployment of convolutional neural networks on mobile devices. In many resource-limited edge devices, complicated operations can be calculated via lookup tables to reduce  computational cost. Motivated by this, in this paper, we introduce a generic and efficient lookup operation which can be used as a basic operation for the construction of neural networks. Instead of calculating the multiplication of weights and activation values, simple yet efficient lookup operations are adopted to compute their responses. To enable end-to-end optimization of the lookup operation, we construct the lookup tables in a differentiable manner and propose several training strategies to promote their convergence. By replacing computationally expensive multiplication operations with our lookup operations, we develop lookup networks for the image classification, image super-resolution, and point cloud classification tasks. It is demonstrated that our lookup networks can benefit from the lookup operations to achieve higher efficiency in terms of energy consumption and inference speed while maintaining competitive performance to vanilla convolutional networks. Extensive experiments show that our lookup networks produce state-of-the-art performance on different tasks (both classification and regression tasks) and different data types (both images and point clouds).
		\end{abstract}
		
		}

	\maketitle

	\IEEEdisplaynontitleabstractindextext

	%
	\IEEEpeerreviewmaketitle

	\IEEEraisesectionheading{\section{Introduction}}
	
	\IEEEPARstart{T}{he} last decade has witnessed the huge success of deep learning since AlexNet \cite{Krizhevsky2012Imagenet} won the champion of the 2012 ImageNet Large Scale Visual Recognition Challenge. With recent advances in deep learning, deep neural networks have achieved remarkable performance in computer vision \cite{He2016Deep,He2017Mask,Chen2018DeepLab,Guo2022Soft}, natural language processing \cite{Parikh2016decomposable,Vaswani2017Attention,Devlin2018Bert}, and many other fields \cite{Schrittwieser2020Mastering,Jumper2021Highly}. However, these networks produce promising results at the cost of high computational complexity and energy consumption, which hinders their deployment on mobile devices. 
	
	To reduce the computational complexity of neural networks, many efforts have been made to develop efficient network architectures, such as SqueezeNets \cite{Iandola2016SqueezeNet}, MobileNets \cite{Howard2017Mobilenets,Sandler2018Mobilenetv2}, and ShuffleNets \cite{Zhang2018Shufflenet,Ma2018Shufflenet}. Different from these hand-crafted networks, neural architecture search (NAS) becomes increasingly popular in designing efficient networks, with MnasNet \cite{Tan2019Mnasnet}, EfficientNet \cite{Tan2019EfficientNet}, and FBNets \cite{Wu2019Fbnet,Wan2020Fbnetv2} being developed. By extensively tuning the width, depth, and convolution types, the networks designed by NAS achieves a better trade-off between accuracy and efficiency than manually developed ones.
	
	In addition to lightweight network architecture designs, a range of generic network acceleration techniques have been widely studied, including weight decomposition \cite{Novikov2015Tensorizing,Yu2017compressing}, network pruning \cite{He2018Soft,He2019Filter,Lin2020HRank}, network quantization \cite{Cao2019Seernet,Wang2018Two,Jung2019Learning}, and knowledge distillation \cite{Yim2017Gift,Xu2020Knowledge,Liu2020Structured}. These techniques can be used to improve the inference efficiency of existing networks and are widely applied in real-world applications.
	
	Different from the aforementioned techniques that focus on reducing redundant computation in the network at the level of channels and layers, several efforts are made to achieve efficient inference of networks from the perspective of basic operations. 
	For convolutional neural networks, convolution operation is the basic operation and takes the majority of  computational cost. 
	Within a convolution operation, multiplication is slower and more energy-intensive than addition \cite{Chen2020AdderNet}.
	Consequently, higher efficiency can be achieved if multiplication can be replaced with cheaper operations. To this end, network binarization techniques \cite{Rastegari2016Xnor,Liu2018Bi} are proposed to calculate the multiplication of binarized activations and weights using XNOR operation. Recently, AdderNet \cite{Chen2020AdderNet} is developed to measure the {correlation} between activations and weights using $\ell_1$ distance rather than cross-correlation. Therefore, multiplication operations can be abandoned to achieve higher speedup and lower energy consumption. 
	
	\begin{figure}[t]
		\centering
		\includegraphics[width=1\linewidth]{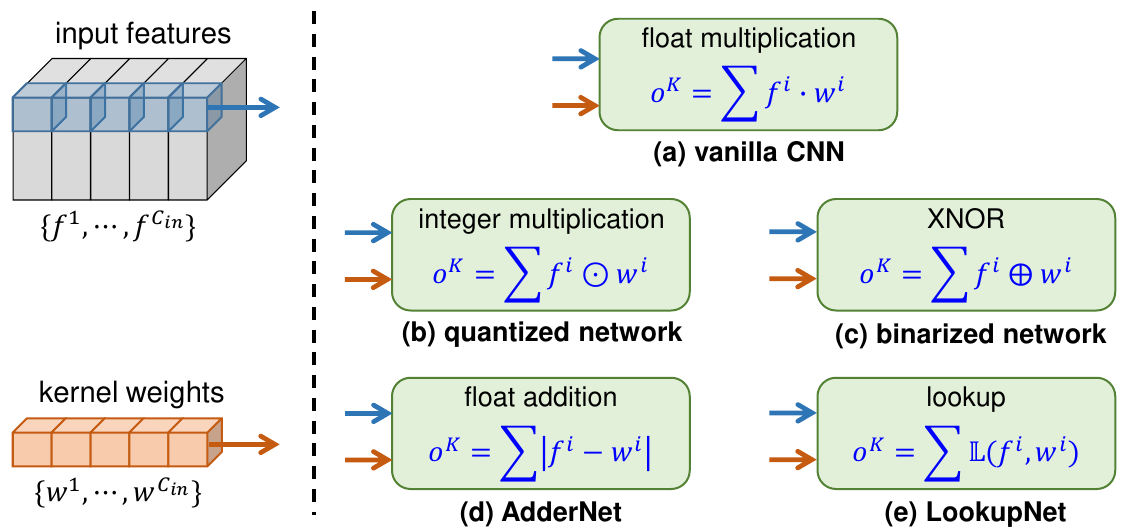}
		\caption{Comparison of basic operations in vanilla convolutional network (a), quantized network (b), binarized network (c), AdderNet (d), and our LookupNet (e).}
		\label{fig0}
	\end{figure}
	
	Intuitively, in a quantized network (\emph{e.g.}, 4-bit), quantized weights and activations have 16 (i.e., $2^{4}$) possible values. Therefore, the multiplication between a weight value and an activation value in a convolution can be simply calculated via a cheap lookup operation over a $16\times16$ table. Moreover, it is demonstrated in \cite{Wang2022Learnable} that network quantization can also be achieved using lookup operation. Consequently, we are motivated to study the feasibility of replacing multiplications with lookup operations to directly map a pair of activation and weight values to their response in neural networks.
	
	In this paper, we develop lookup networks that is built with addition and lookup operations. Instead of calculating the multiplication of weights and activation values, our lookup networks adopt simple yet efficient lookup operations to compute their responses (Fig.~\ref{fig0}(e)). To enable end-to-end optimization of our lookup networks, we construct the lookup tables (LUTs) in a differentiable manner and propose several training strategies to promote their convergence. Benefited from the lookup operations, our lookup networks achieve higher efficiency while producing competitive results as compared to vanilla convolutional networks. 
	
	The contributions of this paper can be summarized as follows:
	\begin{itemize}
		\item We introduce a new type of basic operation, namely lookup operation, that is well compatible to existing operations to construct neural networks. To make this operation differentiable, we develop differentiable lookup tables and propose several training strategies for optimization.
		
		\item We develop a new type of neural network, namely lookup network, that is constructed using lookup and addition operations. Compared to vanilla convolutional networks, our lookup networks remove costly multiplication operations and leverage lookup operations to calculate the response of weights and activations.  
		
		\item We successfully apply our lookup networks to three representative tasks: image classification, image super-resolution (SR), and point cloud classification. Extensive experiments show that our lookup networks achieve state-of-the-art performance on these tasks while achieving superior efficiency as compared to convolutional networks.
	\end{itemize}
	
	This paper is an extension of our previous conference version \cite{Wang2022Learnable}. Compared to the conference version, this paper has a number of significant differences:
	\begin{itemize}
		\item \textbf{Different Objectives:} The conference version aims to improve the inference efficiency of networks via network quantization. In this paper, we aim to develop a generic and efficient lookup operation which can be used as a basic operation to construct efficient neural networks.
		
		\item \textbf{Multiplication-Free:} The conference version uses an 1D lookup table that maps float values to quantized values and multiplication operation is still conducted on  quantized values. In this paper, we develop a 2D lookup table that uses activations and weights as indices to directly obtain their responses from the table without relying on any multiplication operation.
		
		\item \textbf{More Analyses:} We investigate more technical details and provide more analyses with respect to our lookup tables and lookup operations.
	\end{itemize}
	
	The rest of this paper is organized as follows. In Section \ref{Sec2}, we review the related work. In Section \ref{Sec3}, we present our lookup network in details. In Sections \ref{Sec4}, we conduct experiments by applying our lookup networks to the image classification, image SR, and point cloud classification tasks. Finally, we conclude this paper in Section \ref{Sec5}.

	\section{Related Work}
	\label{Sec2}
	
	{In this section, we first briefly review several neural network acceleration techniques that are related to our work, including network quantization and cheap operation design. Then, we discuss related works also with lookup operations.}

	\subsection{Network Quantization}
	
	Network quantization aims at reducing bit-widths of weights and activations in a network for memory and computational efficiency. {Uniform quantization approaches \cite{Zhou2016Incremental,Tung2018Clip,Zhao2019Focused,Esser2019Learned,Yang2020Searching} map full-precision values to uniform quantization levels while non-uniform quantization approaches \cite{Ullrich2017Soft,Zhang2018LQ,Xu2018Deep,Yamamoto2021Learnable} use non-uniform levels to quantize the weights and activations to match their distributions.}
	Since the mapping from float values to discrete quantized values are non-differentiable, straight through estimator (STE) \cite{Bengio2013Estimating} is usually applied to calculate the gradients. Then, Gong \emph{et al.} \cite{Gong2019Differentiable} proposed to represent the quantization function as a concatenation of tanh functions to mitigate gradient approximation errors incurred by STE. Jung \emph{et al.} \cite{Jung2019Learning} used a hand-crafted non-linear function with learnable parameters as the quantization function and used the task loss to optimize these parameters. 
	Yang \emph{et al.} \cite{Yang2019Quantization} shared a similar motivation and used a linear combination of sigmoid functions with learnable biases and scales to represent quantization functions for optimization. 
	Recently, Yang \emph{et al.} \cite{Yang2020Searching} regarded the network quantization as a search of quantized value for each float value and used a differential method for optimization. Zhuang \emph{et al.} \cite{Zhuang2020Training} used an auxiliary module connected to a low-bit network to provide gradients for optimization.

	\begin{figure*}[t]
		\centering
		\includegraphics[width=1\linewidth]{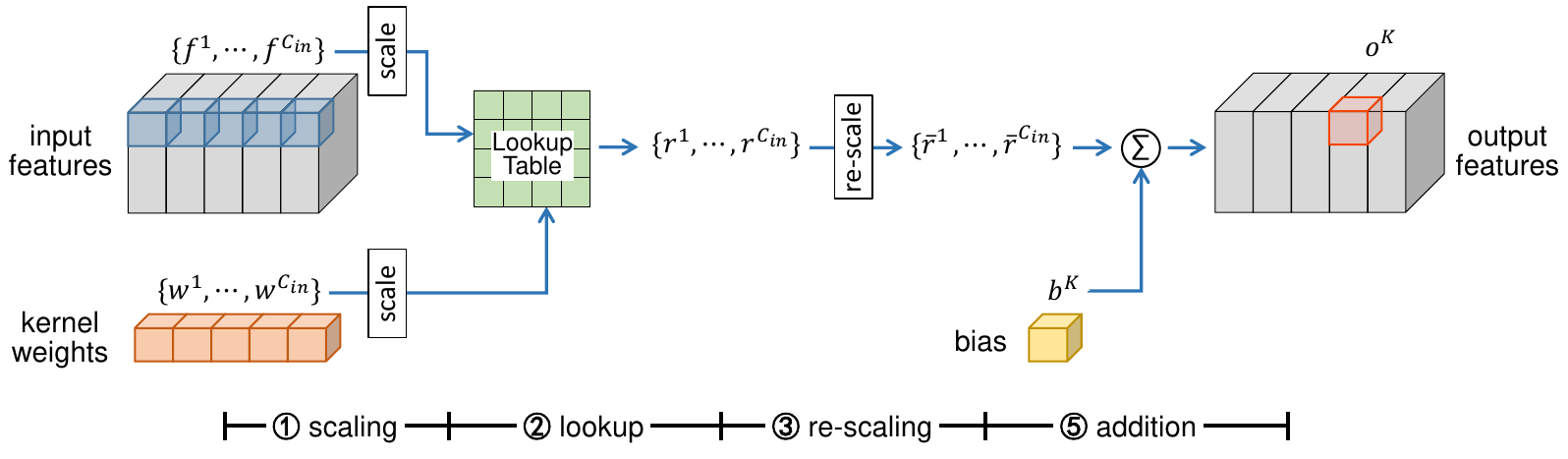}
		\caption{An illustration of our lookup layer. {An $1\times1$ lookup layer is visualized for simplicity. To calculate the response located at $(i,j)$ of the $K^{th}$ channel ($o^{K}$), input features located at $(i,j)$ ($\{f^1,...,f^{C_{in}}\}$) and the kernel weights of the $K^{th}$ output channel ($\{w^1,...,w^{C_{in}}\}$) are scaled to produce indices. Then, these indices are used to find the corresponding responses in the lookup table. Next, the responses are rescaled and summed with an optional bias to produce the final output $o^{K}$.}}
		\label{fig1}
	\end{figure*}
	
	\subsection{Cheap Operation Design}
	
	In addition to the aforementioned filter-level and layer-level techniques, many efforts have been made to investigate cheap operations to achieve network acceleration. 
	\textcolor{black}{Since convolution operation is the most basic operation in a convolutional neural network}, most works focus on replacing it with cheaper ones. {Specifically, Courbariaux \emph{et al.} \cite{Courbariaux2015BinaryConnect} and Zhu \emph{et al.} \cite{Zhu2017Trained} used binary $\{+1,-1\}$ and ternary values $\{+1,0,-1\}$ to represent weight values and realized convolution between weights and activations using only addition operations.
	Rastegari \emph{et al.} \cite{Rastegari2016Xnor} and Lin \emph{et al.} \cite{Lin2017Towards} further represented weight and activation values with binary values $\{-1,+1\}$ such that costly convolution can be achieved using simple XNOR operation.} Chen \emph{et al.} \cite{Chen2020AdderNet} developed AdderNet by abandoning multiplication operations and maximizing the use of addition operations. 
	To increase the receptive field, spatial convolutions (\emph{e.g.}, $3\times3$ convolutions) are widely applied in convolutional networks. To decrease the computational cost that grows quadratically to the kernel size, Wu \emph{et al.} \cite{Wu2018Shift} combined spatial shift operation with point-wise convolutions to achieve $3\times3$ receptive fields while maintaining high efficiency. 
	Recently, Elhoushi \emph{et al.} \cite{Elhoushi2021Deepshift} constructed DeepShift models by using bitwise shift and sign flipping operations to replace the costly multiplication operation in the convolution. Inspired by the success of AdderNet and DeepShift, You \emph{et al.} \cite{You2020Shiftaddnet} further combined complementary bitwise shift and add operations to develop an energy-efficient ShiftAddNet.d
	
	\subsection{CNNs with Lookup Tables}
	
	{As a highly hardware-friendly operation, lookup operation has been studied in several existing approaches. To reduce the redundancy in convolutional kernels, Bagherinezhad \emph{et al.} \cite{Bagherinezhad2017Lcnn} constructed a filter dictionary and employed lookup operation to select a few filters to formulate the kernel. However, multiplication between the weight and activation values is still required. Later, Wang \emph{et al.} \cite{Wang2020LUTNet} developed an area-efficient FPGA-based network termed LUTNet that leverages lookup operation to perform K-input Boolean operation. Nevertheless, the lookup operation takes \textit{K} binary values as input and outputs a binary result, which limits its performance.	Recently, Xu \emph{et al.} \cite{Xu2021Multiplication} constructed a lookup-table-based multiplier to calculate the integer multiplication in quantized networks using lookup operations. Despite superior efficiency, the input and output of the lookup table are constrained to integer values, which hinders further performance improvement.}

	{While network quantization represents the full-precision numbers with low-bit ones for memory and computational efficiency, this paper shares a similar objective with cheap operation designs and aims at replacing expensive operations with cheaper ones. In contrast to existing approaches, we introduce a new lookup operation and combine it with addition operation to develop lookup networks, as shown in Fig.~\ref{fig0}(e). Previous LUT-based approaches commonly use pre-defined lookup tables and limits the inputs/outputs of the table as quantized numbers. In contrast, our lookup table is learnable and directly maps float inputs to a corresponding stored float value with higher capacity. In addition, our method is complementary and compatible to previous network acceleration techniques like network quantization and network pruning for further efficiency improvement (Table~\ref{tab-network}). }

	\section{Methodology}
	\label{Sec3}
	
	Lookup layer is the basic module of our lookup network. It adopts simple lookup and addition operations to calculate the response of the kernel and the input features. In this section, we first present the architecture of the lookup layer. Then, we introduce the construction of lookup table, gradient derivation, and training strategies for the lookup layer. 
	
	\subsection{Architecture}

	In this part, we introduce the architectures of our lookup layer during the training and inference phases.
	
	\subsubsection{Training-Time Architecture}
	As shown in Fig.~\ref{fig1}, our lookup layer takes the features $F$ (\emph{e.g.}, $H\times{W}\times{C_{in}}$), the kernel $w$ (\emph{e.g.}, $C_{out}\times{C_{in}\times1\times1}$), and the bias $b$ (\emph{e.g.}, $C_{out}$) as its input to calculate output features $O$ ($H\times{W}\times{C_{out}}$). During the training phase, our lookup layer consists of four steps, including scaling, lookup, re-scaling, and addition.
	
	\noindent \textbf{(1) Scaling Step} 
	
	To calculate the response located at $(i,j)$ of the $K^{th}$ channel (\emph{i.e.}, $o^{K}$), input features located at $(i,j)$ (\emph{i.e.}, $\{f^1,...,f^{C_{in}}\}$) and the kernel weights of the $K^{th}$ output channel (\emph{i.e.}, $\{w^1,...,w^{C_{in}}\}$) are scaled to produce indices for the lookup table. To this end, $w^1$ and $f^1$ are first normalized using trainable scale parameters $s_w$ and $s_f$. Then, the results are clipped to $[-1,1]$ and $[0,1]$, respectively. Next, the clipped values are discretized to integer indices (\emph{i.e.}, $\{0,...,N_w\!-\!1\}$ and $\{0,...,N_f\!-\!1\}$), respectively. In summary, the indices are calculated as:
	\begin{equation}
		\label{eq1}
		\left\{
		\begin{aligned}
			{idx}_w^c&={\rm discret}\left({\rm clip}\left(\frac{w^c}{s_w}\right);~\left\{0,...,N_w\!-\!1\right\}\right)\\
			{idx}_f^c&={\rm discret}\left({\rm clip}\left(\frac{f^c}{s_f}\right);~\left\{0,...,N_f\!-\!1\right\}\right)
		\end{aligned}
		\right.,
	\end{equation}
	where $c$ is the channel index, $N_{f}$ and $N_{w}$ represent the size of the 2D lookup table.
	
	\noindent \textbf{(2) Lookup Step} 
	
	After the scaling step, the float feature and weight values are mapped to integer indices. Then, these indices are used to find their corresponding response from a lookup table:
	\begin{equation}
		\label{eq2}
		r^c=\mathbb{L}\left({idx}_w^c;~{idx}_f^c;~{\rm T^{\emph l}}\right),
	\end{equation}
	{where ${\rm T^{\emph l}}$ represents a ${{N_{f}\times{N_{w}}}}$ 2D lookup table for the $l^{\rm th}$ layer.} Note that, the values in the lookup table are constrained within $[-1,1]$, which will be detailed in Sec.~\ref{Sec3.2}. 
	
	\noindent \textbf{(3) Re-scaling Step} 
	
	To achieve stable optimization, it is important that that the response $r^c$ has the same magnitude as the input feature $f^c$. Therefore, $r^c$ is re-scaled using scale parameters $s_w$ and $s_f$:
	\begin{equation}
		\label{eq3}
		\overline{r}^c=s_ws_f\times r^c.
	\end{equation}
	
	\noindent \textbf{(4) Addition Step} 
	
	Once response values for different input channels are obtained from the lookup table, $o^{K}$ is finally calculated by accumulating these responses and adding the bias:
	\begin{equation}
		\label{eq4}
		o^{K}=\sum_{c=1}^{C_{in}}\overline{r}^{c}+b^{K}.
	\end{equation}
	
	Note that, a lookup layer with a $1\times1$ kernel and stride of 1 is used for the simplicity of illustration. 
	In practice, our lookup layer can be easily extended to different kernel sizes, stride values, and dilation values, thereby can be used as a basic module to construct efficient neural networks. 
	
	\subsubsection{Inference-Time Architecture}
	\begin{figure*}[t]
		\centering
		\includegraphics[width=1\linewidth]{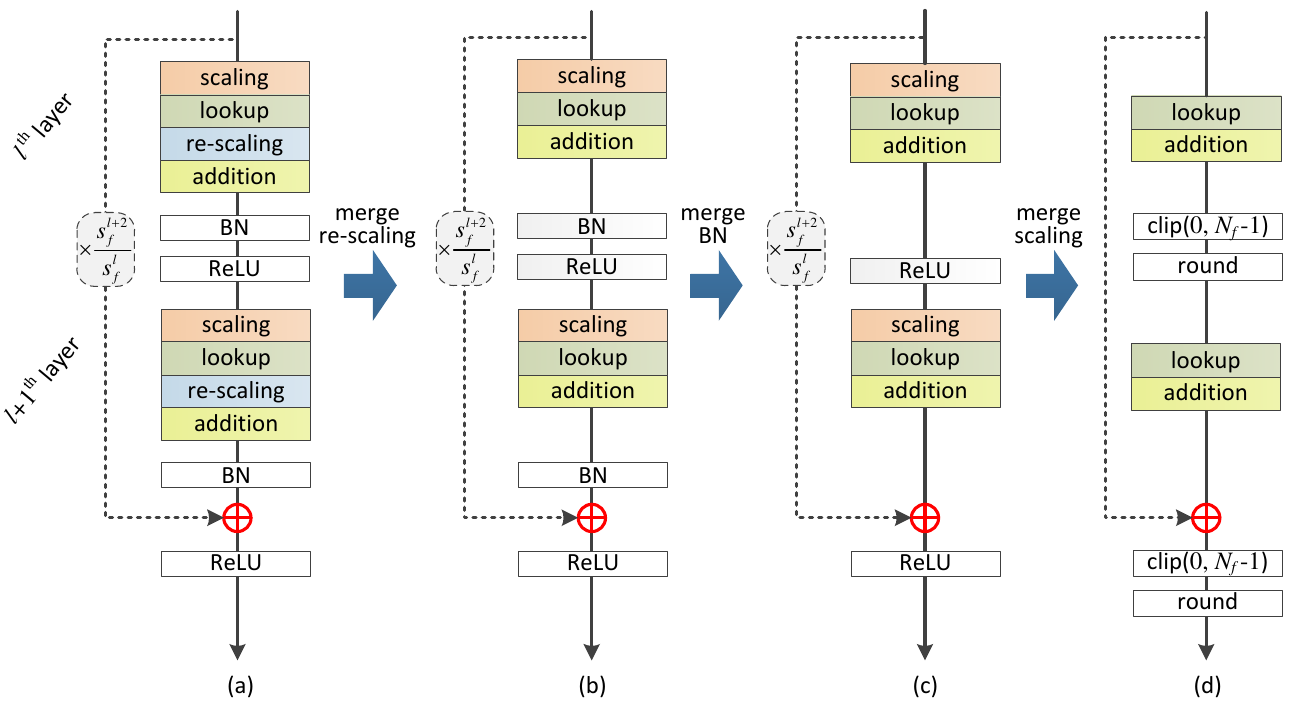}
		\caption{An illustration of our re-parameterization strategy. (a) shows the training-time architecture. (b-d) illustrates the architectures after merging the re-scaling step, the BN layer, and the scaling step, respectively.}
		\label{fig3}
	\end{figure*}
	
	Lookup and addition steps are the key steps in our lookup layer during the training phase. Meanwhile, we still have multiplication operation in the scaling and re-scaling steps. 
	Besides, our lookup layer is usually followed by a BN layer, which also consists of multiplication operations.
	In this part, we introduce a re-parameterization strategy to convert a lookup layer in a trained model to a multiplication-free structure for higher inference efficiency. Specifically, our strategy consists of three steps:
	
	\begin{figure*}[t]
		\centering
		\includegraphics[width=0.97\linewidth]{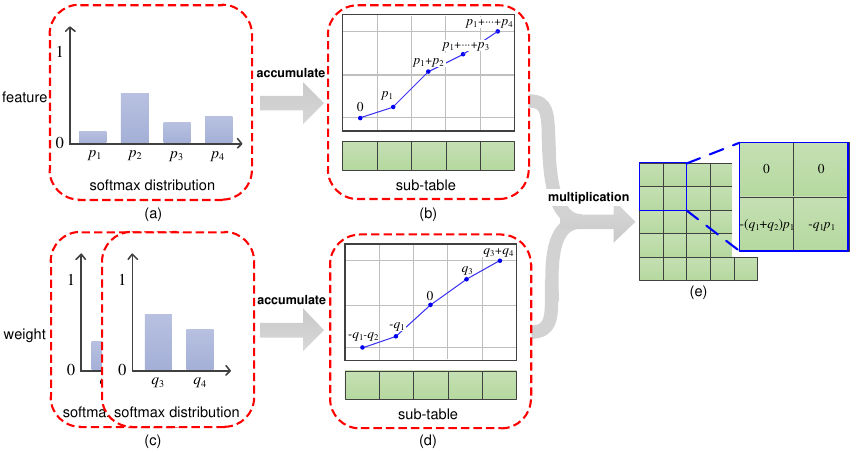}
		\caption{An illustration of the construction of our lookup table. {To obtain a 2D lookup table ($N_f\times N_w$) with monotonicity along both axes at any cell, we decompose it into two 1D sub-tables and formulate each sub-table using cumulative softmax distributions. Particularly, two cumulative softmax distributions are employed to represent the positive and negative axes separately for the sub-table of weights. For simplicity of visualization, $N_f$ and $N_w$ are set to 5.} }
		\label{fig2}
	\end{figure*}

	\noindent\textbf{(1) Merging the Re-scaling Step}
	
	First, as shown in Fig.~\ref{fig3}(b), the re-scaling step is merged into the lookup step by multiplying the lookup table $\rm T$ with scale parameters $s_w$ and $s_f$. As a result, the re-scaled lookup table $s_ws_f{\rm T}$ is directly used for the lookup step and the output of the lookup layer is:
	\begin{equation}
		o^{K}=\sum_{c=1}^{C_{in}}\overline{r}^{c}+b^K=\sum_{c=1}^{C_{in}}\mathbb{L}\left({idx}_w^c;~{idx}_f^c;~s_ws_f{\rm T}\right)+b^K.
	\end{equation}
	Consequently, the multiplication operations at the re-scaling step can be avoided.
	
	\noindent\textbf{(2) Merging the BN Layer}
	
	Second, the BN layer is merged into the lookup step, as illustrated in Fig.~\ref{fig3}(c). Specifically, the output of the BN layer can be re-written as:
	\begin{equation}
		\begin{aligned}
			{\rm BN}(o^{K})&=\gamma\!\times\!\frac{o^{K}-\mu}{{\sigma}}\!+\!\beta\\
			&=\gamma\!\times\!\frac{\sum_{c=1}^{C_{in}}\mathbb{L}\left({idx}_w^c;~{idx}_f^c;~s_ws_f{\rm T}\right)\!+\!b^K\!-\!\mu}{{\sigma}}\!+\!\beta\\
			&={\sum_{c=1}^{C_{in}}\mathbb{L}\left({idx}_w^c;~{idx}_f^c;\frac{\gamma s_ws_f}{{\sigma}}{\rm T}\right)\!+\!\left(\beta\!+\!\frac{\gamma{b^K}\!-\!\gamma\mu}{{\sigma}}\right)},
		\end{aligned}
	\end{equation}
	where $\gamma$ and $\beta$ are learnable parameters, $\mu$ and $\sigma$ are the mean and standard-deviation values for the corresponding channel, respectively. By updating the lookup table to $\frac{\gamma s_ws_f}{{\sigma}}{\rm T}$ and the bias to $\beta\!+\!\frac{\gamma{b^K}-\gamma\mu}{{\sigma}}$, the BN layer can be merged into the lookup layer.
	
	\noindent\textbf{(3) Merging the Scaling Step}
	
	Third, the scaling step is merged into its previous lookup layer to remove its multiplication operations, as shown in Fig.~\ref{fig3}(d). Specifically, for the kernel weights, the scaling step can be executed offline to transform float weights to integer indices without any overhead during inference. For the input features, the lookup table in the previous layer is first multiplied with a scale parameter $\frac{N_f-1}{s_f}$. Then, the ReLU layer in the previous layer is updated to a clip layer (which clip values to $[0,N_f-1]$) and a round layer by incorporating the clip and discretization operations in Eq.~\ref{eq1}. For a multi-branch structure (\emph{e.g.}, residual block), the $(l\!-\!1)^{\rm th}$ layer needs to incorporate the scaling steps in both the $l^{\rm th}$ and the $(l\!+\!2)^{\rm th}$ layers. However, these two layers have different scale parameters (\emph{i.e.}, $s_f^l$ and $s_f^{l+2}$), which leads to a conflict issue. To remedy this, the residual path is multiplied with a scale factor of $\frac{s_f^{l\!+\!2}}{s_f^l}$ during training. As a result, the multiplication operations in the residual path can be avoided at inference time (Fig.~\ref{fig3}(d)).
	
	Overall, by merging the scaling step, the re-scaling step, and the BN layer, our lookup layer removes the multiplication operations and achieves low computational complexity with only lookup and addition operations.
	
	\subsection{Lookup Table Construction}
	\label{Sec3.2}
	
	{Instead of storing pre-computed responses between the indices in the lookup table, we aim to parameterize these responses for joint optimization to adapt them to diverse network architectures and tasks.}
	To achieve stable optimization of the lookup layer, $\mathbb{L}(idx_w^c;idx_f^c;{\rm T})$ in Eq.~\ref{eq2} should be monotonic with respect to $idx_w^c$ and $idx_f^c$. That is, the lookup table $\rm T$ should keep monotonicity along both axes at any cell. To this end, we decompose the 2D lookup table ${\rm T}\in{\mathbb{R}^{N_{f}\times{N_{w}}}}$ along two axes, resulting in two 1D sub-tables ${\rm T}_{f}\in{\mathbb{R}^{N_f\times1}}$ and ${\rm T}_{w}\in{\mathbb{R}^{1\times{N_w}}}$.  
	To ensure the monotonicity of each sub-table, we use the accumulation of a non-negative distribution to construct the sub-table, as illustrated in Fig.~\ref{fig2}. Here, softmax distributions are adopted to make the cumulative distribution function bounded for stable optimization.

	Since feature values after the ReLU layer are non-negative while weight values are not, the construction of these two sub-tables are different. 
	For sub-table of feature values (\emph{e.g.}, ${\rm T}_f\in{\mathbb{R}}^{N_f\times1}$), we first generate a softmax distribution $\{p_1,...,p_{N_{f}-1}\}$ (Fig.~\ref{fig2}(a)):
	\begin{equation}
		p_i=\frac{{\rm exp}(g_i)}{\sum_{i=1}^{N_f-1}{{\rm exp}(g_i)}},
	\end{equation}
	where $g_i$ is a learnable parameter. Then, the softmax distribution is accumulated to construct a sub-table (Fig.~\ref{fig2}(b)).
	For sub-table of weight values (\emph{e.g.}, ${\rm T}_w\in{R}^{1\times N_w}$), we first generate two softmax distribution $\{q_1,...,q_{\frac{N_{w}-1}{2}}\}$ and $\{q_{\frac{N_{w}+1}{2}},...,q_{N_{w}-1}\}$ (Fig.~\ref{fig2}(c)) and then accumulate these two distributions to construct a sub-table (Fig.~\ref{fig2}(d)).
	Next, ${\rm T}_f$ and ${\rm T}_w$ are multiplied to produce the final lookup table $\rm T$ (Fig.~\ref{fig2}(e)). Each entry in this table is the multiplication of corresponding values in these two sub-tables. {Thanks to the softmax distributions, all entries in the table are constrained within $[-1,1]$.} 
	
	\subsection{Gradient Derivation}
	
	In this part, we present the derivation of gradients at different steps in our lookup layer.
	
	\noindent \textbf{(1) Scaling Step} 
	
	During the backward propagation of the scaling step, the gradients of the kernel weights and the scale parameters are derived as:
	\begin{equation}
		\left\{
		\begin{aligned}	
			&\frac{\partial{idx_w^c}}{\partial{w^c}}
			=\left\{
			\begin{array}{lr}
				1/{s_w}, ~~~~~~~~~~~~~~&{\rm if}~w^c<s_w \\
				0, &{\rm otherwise}~~\\
			\end{array}
			\right.
			\\
			&\frac{\partial{idx_w^c}}{\partial{s_w}}
			=\left\{
			\begin{array}{lr}
				-w^c/(s_w)^2, ~~~&{\rm if}~w^c<s_w \\
				0, &{\rm otherwise}~~\\
			\end{array}
			\right.
		\end{aligned}.
		\right.
	\end{equation}
	Note that, the straight-through-estimator (STE) \cite{Bengio2013Estimating} is adopted to compute the gradient for the discretization function in Eq.~\ref{eq1}. The gradients of the input features and the corresponding scale parameters can be derived similarly.
	
	\noindent \textbf{(2) Lookup Step} 
	
	To make our lookup step differentiable, STE \cite{Bengio2013Estimating} is used to calculate gradients during backward propagation. Given indices $idx_w^c$ and $idx_f^c$, assume $r^c=(p_1+p_2)(q_3+q_4)$ is found from the lookup table, the gradients of $idx_w^c$ and the lookup table can be derived as:
	\begin{equation}
		\left\{
		\begin{aligned}	
			&\frac{r^c}{\partial{idx_w^c}}=1\\
			&\frac{r^c}{\partial{p_1}}=q_3+q_4\\
			&\frac{r^c}{\partial{q_3}}=p_1+p_2
		\end{aligned}.
		\right.
	\end{equation}
	The gradients of the input features can be derived similarly.
	
	\noindent \textbf{(3) Re-scaling Step} 
	
	The re-scaling step is naturally differentiable and the gradients can be easily obtained as:
	\begin{equation}
		\left\{
		\begin{aligned}	
			&\frac{\overline{r}^c}{\partial{r^c}}=s_ws_f\\
			&\frac{\overline{r}^c}{\partial{s_w}}=s_f{r}^c\\
			&\frac{\overline{r}^c}{\partial{s_f}}=s_w{r}^c
		\end{aligned}.
		\right.
	\end{equation}
	
	\noindent \textbf{(4) Addition Step} 
	
	The addition step is also differentiable and its gradients can be easily derived:
	\begin{equation}
		\frac{\partial{o^K}}{\partial{\overline{r}^c}}=1.
	\end{equation}
	
	In summary, our lookup layer is fully differentiable and can be optimized with other modules within a neural network in an end-to-end manner.
	
	\begin{figure*}[t]
		\centering
		\includegraphics[width=0.88\linewidth]{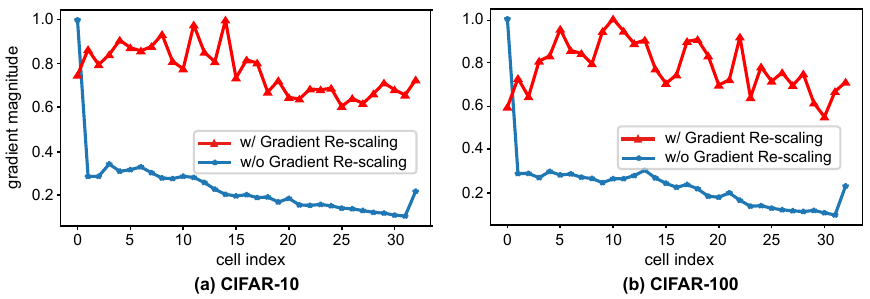}
		\caption{An illustration of gradient imbalance among different cells of our sub-table. Normalized gradient magnitudes averaged over 100 iterations are shown.}
		\label{fig4}
	\end{figure*}
	
	\subsection{Training Strategies}
	\label{sec3.4}
	
	To promote the convergence of our lookup layer, we introduce two training strategies.
	
	\noindent \textbf{(1) Exponential Formulation of Scale Parameter}
	
	The scale parameters $s_w$ and $s_f$ in Eq.~\ref{eq1} are positive values used to normalize kernel weights and input features. In our experiments, it is observed that these scale parameters may become negative during training, which leads to convergence issues. To address this problem, we introduce auxiliary parameters $e_w$ and $e_a$ to formulate scale parameters as:
	\begin{equation}
		\label{eq12}
		\left\{
		\begin{array}{lr}
			s_w={\rm exp}(e_w)\\
			s_a={\rm exp}(e_a)\\
		\end{array}
		\right..
	\end{equation}
	During training, $e_w$ is initialized using the standard deviation of weights ($\ln(3\sigma_w)$) and $e_a$ is initialized using the standard deviation of feature values in the first iteration ($\ln(3\sigma_f)$).

	\noindent \textbf{(2) Gradient Re-scaling}
	
	In our experiments, we observe gradient imbalance among different cells of our lookup table.
	Due to the bell-shaped distributions of features and kernel weights, the numbers of values falling into different cells during the lookup operation are quite different. Consequently, aggregated gradients of cells near 0 are much larger than those near 1 (blue curve in Fig.~\ref{fig4}) and dominate the optimization of the lookup table. To handle this problem, gradient $g_i$ of the $i^{\rm th}$ cell is re-scaled using $\sqrt\frac{{N_{avg}}}{{{N_i}}}$, where $N_{avg}$ is the average number of float values in a cell and $N_i$ is the number of float values falling in the $i^{\rm th}$ cell. With our gradient re-scaling scheme, gradients over different cells of the lookup table are balanced, as shown in Fig.~\ref{fig4}. 
	
	\begin{table*}[t]
		\caption{Top-1 accuracy achieved on CIFAR-10 and CIFAR-100 with different settings. {Models 1-7 are model variants with different settings, which are detailed in Sec.~\ref{sec4.1.1}, \ref{sec4.1.2}, and \ref{sec4.1.4}.} \textcolor{black}{``Granularity'' represents the size of the lookup table (\emph{i.e.,} $N_f\times N_w$). ``Memory'' represents the additional memory consumption to save the lookup table, which is presented in brackets.} ResNet-20 is used as the baseline model.}
		\label{tab-ablation}
		\centering
		\renewcommand\arraystretch{1.2}
		\setlength{\tabcolsep}{2mm}{
			\begin{tabular}{ccccccccccc}
				\toprule[0.75pt] 
				\multirow{2}{*}{Model} & \multicolumn{4}{c}{Lookup Table Construction} & \multirow{3}{*}{\tabincell{c}{Granularity\\(Memory)}} & \multicolumn{2}{c}{Strategy} & {\multirow{3}{*}{\tabincell{c}{CIFAR-10\\(\%)}}} & {\multirow{3}{*}{\tabincell{c}{CIFAR-100\\(\%)}}}
				\tabularnewline
				\cmidrule(lr){2-5}\cmidrule(lr){7-8}
				& Fixed & \tabincell{c}{Independent\\(Random Init.)} & \tabincell{c}{Independent\\(Step Init.)} & Cumulative & & Exponential & Re-scaling
				\tabularnewline
				\hline
				{Baseline} & - & - & - & - & - & - & - & {92.25} & {68.14}
				\tabularnewline
				\hline
				{Model 1} & \ding{51} & & & & $33\times33$ (4.2KB) & - & - & 91.80 & 67.77
				\tabularnewline
				{Model 2} &  & \ding{51} & & & $33\times33$ (4.2KB) & \ding{51} & \ding{51} & 19.17 & 8.16
				\tabularnewline
				{Model 3} &  & & \ding{51} & & $33\times33$ (4.2KB) & \ding{51} & \ding{51} & 43.52 & 25.24
				\tabularnewline
				\hline
				{Model 4} & & & & \ding{51} & $17\times17$ (1.1KB)  & \ding{51} & \ding{51} & 92.21 & 67.85
				\tabularnewline
				{Model 5} & & & & \ding{51} & $65\times65$ (16.5KB) & \ding{51} & \ding{51} & 92.70 & 68.97
				\tabularnewline
				\hline
				{Model 6} & & & & \ding{51} & $33\times33$ (4.2KB) & \ding{55} & \ding{55} & 92.35 & 68.12
				\tabularnewline
				{Model 7} & & & & \ding{51} & $33\times33$ (4.2KB) & \ding{51} & \ding{55} & 92.47 & 68.51
				\tabularnewline
				\hline
				Ours & & & & \ding{51} & $33\times33$ (4.2KB) & \ding{51} & \ding{51}  & 92.68 & 68.96
				\tabularnewline
				\toprule[0.75pt]
		\end{tabular}}
	\end{table*}

	\subsection{Discussion}
	
	In essence, the output features of a basic layer (\emph{e.g.}, a convolutional layer) in a neural network indicate the {correlation} between the input features (\emph{e.g.}, $\{f^1,...,f^{C_{in}}\}$) and the kernel weights (\emph{e.g.}, $\{w^1,...,w^{C_{in}}\}$) \cite{Chen2020AdderNet}:
	\begin{equation}
		o^K=\sum_{c=1}^{C_{in}}S(f^c,~w^c),
	\end{equation}
	where $S(\cdot,\cdot)$ represents a {correlation} measure. Convolutional networks use multiplication (\emph{i.e.}, $S(x,y)=x\times y$) to measure the {correlation} while recent AdderNets adopt an $\ell_1$ distance (\emph{i.e.}, $S(x,y)=|x-y|$) as the metric. 
	
	Different from these networks that use hand-crafted {correlation} measure, our lookup layer employs a learnable lookup table (\emph{i.e.}, $S(x,y)=\mathbb{L}(x,y,{\rm T})$) to obtain the {correlation} between the input features and kernel weights. \textbf{First}, a lookup table with infinite cells can be considered as a general formulation of {correlation} measure. For example, the lookup table can be transformed to a multiplication measure or an $\ell_1$ distance measure when corresponding responses are filled. \textbf{Second}, instead of using fixed and manually defined measure, our lookup table is trainable and can learn {correlation} measure to fit different tasks and data. \textbf{Third}, thanks to the simplicity of the lookup operation, our lookup layer can calculate the {correlation} between the input features and the kernel weights with high efficiency. Particularly, our lookup operation is highly friendly to hardwares like field programmable gate array (FPGA) for more convenient and efficient deployment.

	

	\section{Experiments}
	\label{Sec4}
	
	In this section, we conduct experiments on three representative tasks to validate the effectiveness of our lookup network, including image classification, image SR, and point cloud classification. For fair comparison with previous convolutional networks, we replace their convolutional layers with our lookup layers to construct lookup networks with the same structure.

	\subsection{Model Analyses}
	
	In this part, we conduct experiments on the CIFAR-10 and CIFAR-100 datasets to investigate the effectiveness of our network designs. 
	
	\subsubsection{Lookup Table Construction}
	\label{sec4.1.1}
	
	We conduct experiments to investigate the construction of lookup tables in our network. First, we developed a network variant (model 1) by replacing our learnable lookup tables with fixed ones. More specifically, the lookup tables in this model remain untouched during training. Second, to demonstrate the effectiveness of our lookup table construction using cumulative softmax distributions (Fig.~\ref{fig2}), we developed another two network variants (model 2 and model 3) by constructing lookup tables with independent parameters. For model 2, random values drawn from a uniform distribution $\mathbb{U}(0,1)$ were used for initialization. For model 3, we initialized the lookup tables using values from a step function (\emph{e.g.}, $0,\frac{1}{10},...,\frac{9}{10},1$). We compare the performance of models 1-3 to our network in Table~\ref{tab-ablation}. 
	
	When fixed lookup tables are adopted in the network, model 1 suffers relatively low accuracy. Compared to fixed lookup tables, learnable lookup tables facilitate our network to produce much better performance, with accuracy being improved from 91.80/67.77 to 92.68/68.96. 
	This demonstrates that optimizing the lookup tables together with the network is beneficial to performance improvement. 
	However, when learnable lookup tables are constructed using independent parameters, model 2 and model 3 suffer a severe performance drop with very low accuracy. This is because, 2D lookup tables parameterized with independent values cannot keep monotonicity along two axes, which hinders their optimization during training. By constructing lookup tables using cumulative softmax distributions, 
	better convergence can be achieved such that much higher accuracy can be produced.
	This clearly demonstrates the effectiveness of our lookup table construction. 
	
	\begin{figure}[t]
		\centering
		\includegraphics[width=1\linewidth]{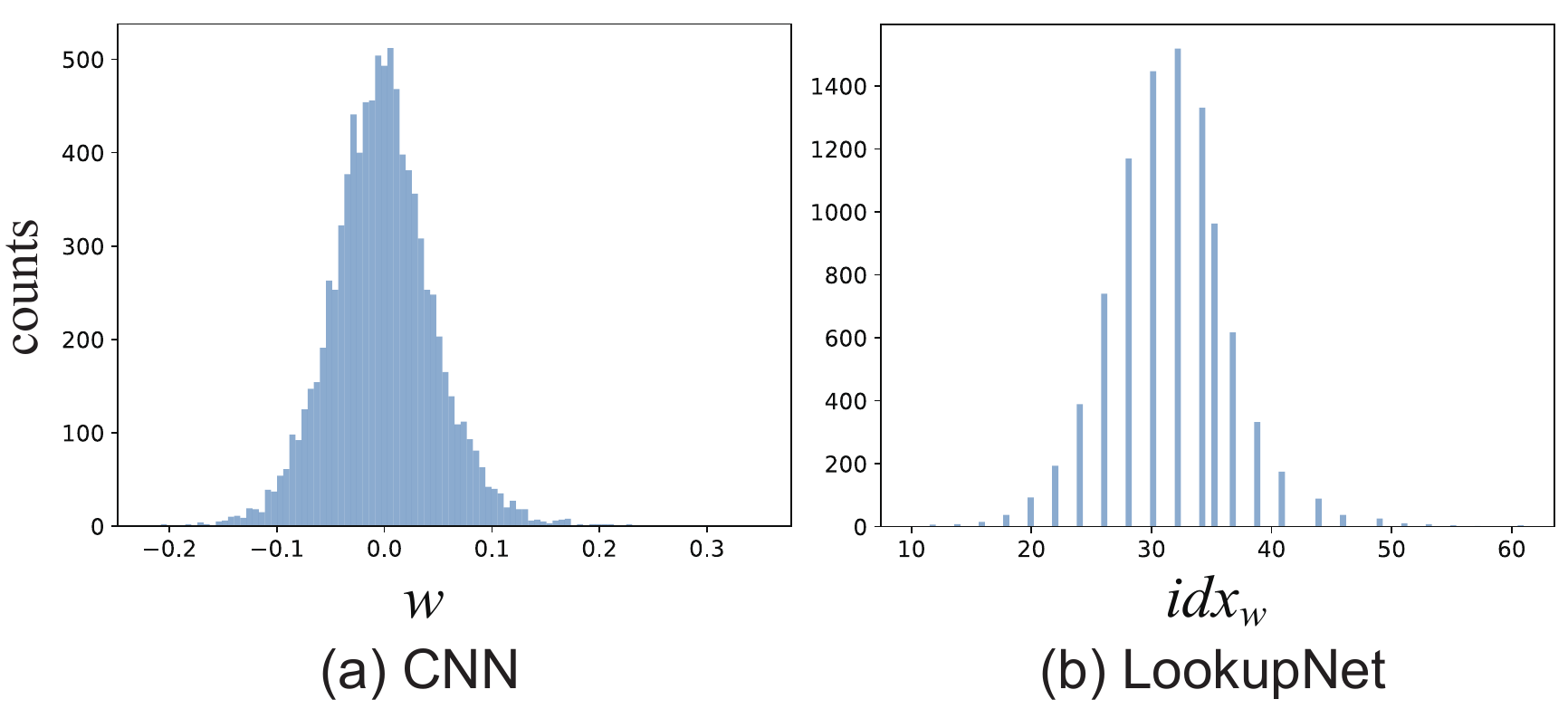}
		\caption{Distributions of the weight values in the vanilla convolutional network (a) and our lookup network (b). Note that, the weight values in the lookup network are integer indices.}
		\label{fig_distribution}
	\end{figure}
	
	\begin{figure*}[t]
		\centering
		\includegraphics[width=0.92\linewidth]{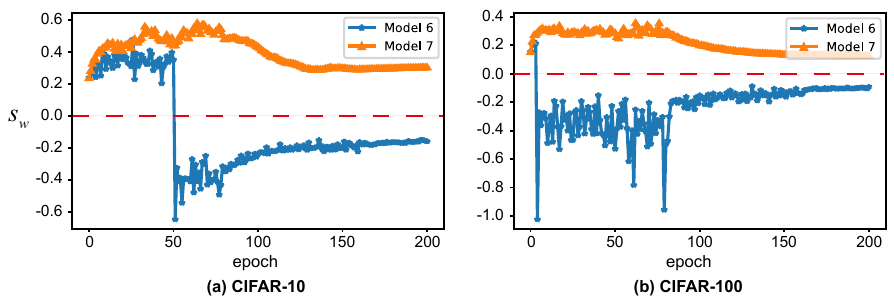}
		\caption{Evolution of scale parameter $s_w$ in model 6 and model 7 during training. {The configurations of these two models are presented in Table~\ref{tab1} and detailed in Sec.~\ref{sec4.1.4}.}}
		\label{fig5}
	\end{figure*}
	
	\subsubsection{Granularity of Lookup Tables}
	\label{sec4.1.2}
	
	The granularity of lookup tables (\emph{i.e.}, $N_f$ and $N_w$ in Fig.~\ref{fig2}) determines their sizes and the degree of freedom during optimization. Intuitively, finer granularity (\emph{i.e.}, larger values for $N_f$ and $N_w$) helps to produce better performance at the cost of higher memory cost. We conduct experiments to study the effect of different granularities. Specifically, we develop two network variants (model 4 and model 5) using lookup tables with different granularities. 
	
	Table~\ref{tab-ablation} compares the accuracy of networks using lookup tables with different granularities. When the granularity is increased from 17 to 33, our network has an accuracy improvement from 92.21/67.85 to 92.68/68.96 on CIFAR-10/CIFAR-100. With larger granularity, the degree of freedom for our lookup tables during optimization is increased such that better performance is achieved. However, further increasing the granularity from 33 to 65 (model 5) only introduces marginal improvements (92.68/68.96 vs. 92.70/68.97) with a $4\times$ memory consumption. Consequently, granularity is set to 33 by default in our networks for a balance between accuracy and efficiency.
	
	\subsubsection{Distributions of Weights}
	\label{sec4.1.3}
	
	We visualize the distributions of weight values for the $9^{\rm th}$ layer in the vanilla convolutional network and our lookup network. As shown in Fig.~\ref{fig_distribution}, the distribution of weights in the convolutional network looks like a Gaussian distribution. In contrast, the weight values in our lookup network are a group of discrete indices which are then used to find corresponding values in the lookup table. In addition, the intervals between these indices are not identical and can adapt to the distribution of weights, which is similar to non-uniform network quantization. However, different from non-uniform network quantization methods that rely on delicate hardware for acceleration \cite{Gong2019Differentiable}, our lookup network can benefit from the efficient lookup operation to achieve practical speedup on general hardwares (as discussed in Sec.~\ref{Sec4.1.6}).

	\subsubsection{Training Strategies}
	\label{sec4.1.4}
	
	\noindent \textbf{(1) Exponential Formulation of Scale Parameter}
	
	Since scale parameters are vulnerable to sign reversal, an exponential formulation is introduced for stable convergence. To demonstrate its effectiveness, we developed a network variant (model 7 in Table~\ref{tab-ablation}) by replacing the scale parameters in model 6 with an exponential formulation (Eq.~\ref{eq12}). Since the sign reversal of scale parameters largely affects the convergence of the network, model 6 suffers relatively low accuracy (92.35/68.12). With our exponential formulation, a good convergence can be achieved such that better performance can be obtained by model 7 (92.47/68.51). 
	
	{Fig.~\ref{fig5}} further plots the curves of scale parameters in models 6 and 7 during training. It can be observed that the scale parameter in model 6 has a violent fluctuation and encounters sign reversals at epoch 50 on CIFAR-10 and epoch 5 on CIFAR-100. Due to the convergence issue caused by the sign reversal, model 6 suffers limited accuracy. With our exponential formulation, the training of scale parameters in model 7 is more stable such that better performance can be obtained.
	
	\begin{figure*}[t]
		\centering
		\includegraphics[width=1\linewidth]{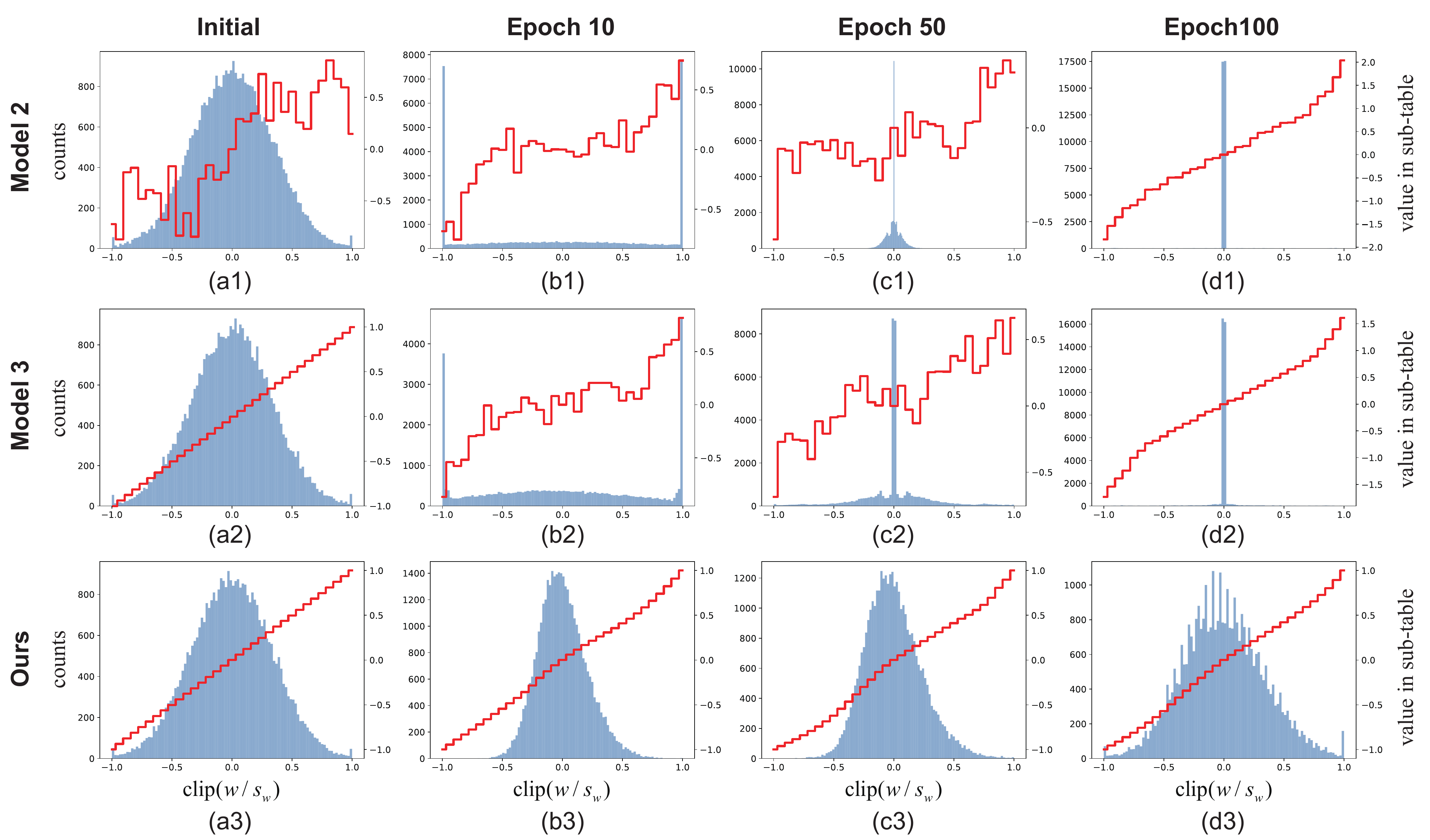}
		\caption{Evolution of lookup tables (red lines) and the distributions  of normalized weights (blue bars) during training on the CIFAR-100 dataset. For simplicity of illustration, only the sub-table for weights ($\mathbb{R}^{1\times{N_w}}$) is shown. {The configurations of model 2 and model 3 are presented in Table~\ref{tab1} and detailed in Sec.~\ref{sec4.1.1}.}}
		\label{fig_evolution}
	\end{figure*}
	
	\noindent \textbf{(2) Gradient Re-scaling}
	
	Gradient re-scaling scheme is introduced to handle the gradient imbalance among different cells of our lookup tables. To demonstrate its effectiveness, we compare the performance of model 7 to our baseline. 
	It can be observed from Table~\ref{tab-ablation} that our gradient re-scaling scheme facilitates our network to obtain higher accuracy than model 7 (92.68/68.96 vs. 92.47/68.51). Without the gradient re-scaling scheme, the gradient imbalance among different cells of our lookup tables hinders a good convergence. As a result, the performance of model 7 is limited. With our re-scaling scheme, gradients over different cells of our lookup tables can be balanced (Fig.~\ref{fig3}) such that a good convergence can be achieved for superior performance.
	
	
	Overall, with both exponential formulation of scale parameters and gradient re-scaling scheme, our network achieves the best performance.

	\begin{table*}[t]
		\setcounter{table}{2}
		\caption{Energy consumption, latency, and top-1 accuracy achieved on CIFAR-10 and CIFAR-100. ResNet-20 is used as the baseline model. {Models 8-11 are model variants with different combinations of model acceleration approaches.}}
		\label{tab-network}
		\centering
		\renewcommand\arraystretch{1.2}
		\setlength{\tabcolsep}{1.4mm}{
			\begin{tabular}{lcccccccccc}
				\toprule[0.75pt] 
				\multirow{2}{*}{Model} & \multicolumn{3}{c}{Approach} & \multirow{2}{*}{Model Size} & \multicolumn{2}{c}{Energy (mJ)} & \multicolumn{2}{c}{Latency (cycle)} & \multicolumn{2}{c}{Acc. (\%)}
				\tabularnewline
				\cmidrule(lr){2-4}\cmidrule(lr){6-7}\cmidrule(lr){8-9}\cmidrule(lr){10-11}
				& Pruning & Quantization & Cheap & & Cortex-A7 & Cortex-A15 & Cortex-A7 & Cortex-A15 & CIFAR-10 & CIFAR-100
				\tabularnewline
				\hline
				Baseline & - & - & - & 1143KB & 15.7 & 124.2 & 312M & 390M & 92.25 & 68.14 
				\tabularnewline				
				\hline
				{Model 8} & \ding{51}\cite{Li2020DHP} &  &  & 502KB & 7.6 & 60.5 & 152M & 190M & 91.54 & -
				\tabularnewline
				\hline
				{Model 9} &  & \ding{51}\cite{Choi2018Pact} &  & 147KB & 8.9 & 49.8 & 156M & 156M & 91.30 & -
				\tabularnewline
				\hline
				{Model 10} & & & XNOR \cite{Zhou2016Dorefa} & 40KB & 10.6 & 72.7 & 195M & 234M & 84.87 & 54.14
				\tabularnewline
				{Model 11} & &  & Add. \cite{Chen2020AdderNet} & 1143KB & 15.5 & 114.7 & 312M & 390M & 91.84 & 67.60
				\tabularnewline
				\hline
				Ours &  &  & Lookup & 1215KB & 13.6 & 75.0 & 195M & 234M & {\textbf{92.68}} & {\textbf{68.97}}
				\tabularnewline
				Ours+pruned & \ding{51}\cite{Li2020DHP} & & Lookup & 538KB & \textbf{6.6} & 36.0 & 95M & 114M & 92.55 & 68.58
				\tabularnewline
				Ours+4bit & & \ding{51}\cite{Choi2018Pact} & Lookup & 155KB & 9.0 & \textbf{34.5} & \textbf{78M} & \textbf{78M} & {92.51} & {68.35}
				\tabularnewline
				\toprule[0.75pt] 
		\end{tabular}}
	\end{table*}	
	
	\begin{table}[t]
		\setcounter{table}{1}
		\caption{Energy consumption and latency per operation at 1G Hz for Cortex-A7, Cortex-A15 processors.}
		\label{tab-ops}
		\centering
		\renewcommand\arraystretch{1.2}
		\setlength{\tabcolsep}{0.9mm}{
			\begin{tabular}{llccccccc}
				\toprule[0.75pt] 
				& & \tabincell{c}{Float\\Add.} & \tabincell{c}{Float\\Mul.} & \tabincell{c}{4-Bit\\Add.} & \tabincell{c}{4-Bit\\Mul.} & XNOR & Shift & Lookup
				\tabularnewline
				\hline
				\multirow{2}{*}{\tabincell{c}{Energy\\(pJ)}}& Cortex-A7 & 199 & 203 & 82 & 146 & 72 & - & 150
				\tabularnewline
				& Cortex-A15 & 1471 & 1714 & 432 & 846 & 394 & - & 452
				\tabularnewline
				\toprule[0.75pt] 
				\multirow{2}{*}{\tabincell{c}{Latency\\(cycle)}}& Cortex-A7 & 4 & 4 & 1 & 3 & 1 & 1 & 1
				\tabularnewline
				& Cortex-A15 & 5 & 5 & 1 & 3 & 1 & 1 & 1
				\tabularnewline
				\toprule[0.75pt] 
		\end{tabular}}
	\end{table}

	\subsubsection{Evolution of Network}
	\label{sec_evolution}
	
	As analyzed above, a good convergence of the network is critical to its performance. Therefore, we illustrate the lookup tables and the distributions of normalized weights (\emph{i.e.}, ${\rm clip}(\frac{w}{s_w})$ in Eq.~\ref{eq1}) at different epochs in Fig.~\ref{fig_evolution} to study their evolution. 
	
	For model 2, we can see that its initial lookup table is not monotonic (Fig.~\ref{fig_evolution}(a1)) since it is initialized using independent random values. Consequently, the optimization of weights and lookup tables is difficult. At the beginning (Fig.~\ref{fig_evolution}(b1)), the distribution of weights have a drastic change with a large quantity of values being clipped (larger than 1 or smaller than -1). As the training continues, the weight values gradually gather around 0, as shown in Fig.~\ref{fig_evolution}(c1). Finally, as illustrated in Fig.~\ref{fig_evolution}(d1), the majority of weights are close to 0 with very small values. As a result, model 2 suffers a low performance of 8.16\%. For model 3, although its lookup table is initialized using values drawn from a step function to achieve the monotonicity (Fig.~\ref{fig_evolution}(a2)), its different cells are still independent. Therefore, the randomness during network training also makes the optimization of weights and lookup tables unstable (Fig.~\ref{fig_evolution}(b2) and (c2)). After convergence, the majority of weights are close to 0 (Fig.~\ref{fig_evolution}(d2)) with limited accuracy being produced (25.24\%). 
	
	In contrast to model 2 and model 3, the lookup table in our network is constructed using a cumulative softmax distribution. From Fig.~\ref{fig_evolution}(a3)-(c3) we can see that, our lookup table keeps the monotonicity during training and contributes to a stable optimization of the network. Besides, our lookup table is trainable and can flexibly adapt to the distribution of the weights during optimization. As illustrated in Fig.~\ref{fig_evolution}(d3), our network reaches a good convergence of lookup table and weights with a high accuracy (68.96\%).



	\subsubsection{Efficiency Evaluation} 
	\label{Sec4.1.6}
	
	\textbf{(1) Theoretical Analyses}
	
	We conduct experiments to study the superior efficiency of our lookup operation. Three families of network acceleration methods are used for comparison, including network pruning based methods, network quantization based methods, and cheap operation based methods. Specifically, four network variants (models 8-11) were developed by applying different techniques \cite{Li2020DHP,Choi2018Pact,Zhou2016Dorefa,Chen2020AdderNet}.  In our experiments, two popular general mobile processors (ARM Cortex-A7 and Cortex-A15) are used for theoretical analyses. 
	Basic energy consumption and latency per operation on these two processors \cite{Vasilakis2015instruction} are shown in Table~\ref{tab-ops}. Comparative results achieved by different models are presented in Table~\ref{tab-network}.

	\begin{table*}[t]
		\setcounter{table}{3}
		\caption{\textcolor{black}{Computational resource consumption (memory, LUT, register, and energy) and running time achieved on different types of hardwares. Results are averaged over 10 runs. ``$B$'' denotes batch size.}}
		\label{tab-time}
		\centering
		\renewcommand\arraystretch{1.2}
		\setlength{\tabcolsep}{1.7mm}{
			\begin{tabular}{llccccccccccccccc}
				\toprule[0.75pt] 
				& & \multicolumn{2}{c}{GPU ($B$=8)} & \multicolumn{2}{c}{GPU ($B$=512)} & \multicolumn{1}{c}{Mobile ($B$=8)} & \multicolumn{3}{c}{FPGA ($B$=8)} & \multicolumn{2}{c}{Acc. (\%)}
				\tabularnewline
				\cmidrule(lr){3-4}\cmidrule(lr){5-6}\cmidrule(lr){7-7}\cmidrule(lr){8-10}\cmidrule(lr){11-12}
				& & Memory & Time & Memory & Time & Time & LUT & Register & Energy & CIFAR-10 & CIFAR-100
				\tabularnewline
				\hline
				\multirow{5}{*}{ResNet-20}
				& Baseline & $1\times$ & $1\times$ & $1\times$ & $1\times$ & $1\times$ & $1\times$ & $1\times$ & $1\times$ & 92.25 & 68.14
				\tabularnewline
				\cline{2-12}
				& Pruning \cite{Li2020DHP} & $\textbf{0.98}\times$ & $0.93\times$ & $\textbf{0.92}\times$ & $0.87\times$ & $0.91\times$ & $0.49\times$ & $0.49\times$ & $0.49\times$ & 91.51 & -
				\tabularnewline
				& AdderNet \cite{Chen2020AdderNet} & $1.10\times$ & $1.03\times$ & $1.03\times$ & $0.96\times$ & $0.96\times$ & $0.46\times$ & $0.67\times$ & $0.62\times$ & 91.84 & 67.60
				\tabularnewline
				& Ours & $1.01\times$ & $0.96\times$ & $1.02\times$ & $\textbf{0.85}\times$ & $0.88\times$ & $0.33\times$ & $0.33\times$ & $0.41\times$ & \textbf{92.68} & \textbf{68.97}
				\tabularnewline
				& Ours-pruned & $0.99\times$ & $\textbf{0.91}\times$ & $0.99\times$ & $0.94\times$ & $\textbf{0.83}\times$ & $\textbf{0.18}\times$ & $\textbf{0.19}\times$ & $\textbf{0.23}\times$ & 92.55 & 68.58 
				\tabularnewline
				\hline
				\multirow{5}{*}{VGG-Small}
				& Baseline & $1\times$ & $1\times$ & $1\times$ & $1\times$ & $1\times$ & $1\times$ & $1\times$ & $1\times$ & 93.80 & 72.73
				\tabularnewline
				\cline{2-12}
				& Pruning \cite{Lin2019Towards} & $0.83\times$ & $0.92\times$ & $0.83\times$ & $0.76\times$ & $0.88\times$ & $0.49\times$ & $0.49\times$ & $0.49\times$ & 93.42 & -
				\tabularnewline
				& AdderNet \cite{Chen2020AdderNet} & $1.06\times$ & $1.04\times$  & $0.98\times$ & $0.94\times$ & $0.98\times$ & $0.46\times$ & $0.67\times$ & $0.62\times$ & 93.72 & 72.64
				\tabularnewline
				& Ours & $0.63\times$ & $0.82\times$ & $0.67\times$ & $0.56\times$  & $0.83\times$ & $0.33\times$ & $0.33\times$ & $0.41\times$ & \textbf{94.14} & \textbf{75.75} 
				\tabularnewline
				& Ours-pruned & $\textbf{0.57}\times$ & $\textbf{0.78}\times$ & $\textbf{0.60}\times$ & $\textbf{0.50}\times$ & $\textbf{0.77}\times$ & $\textbf{0.18}\times$ & $\textbf{0.19}\times$ & $\textbf{0.23}\times$ & 94.02 & 75.26 & 
				\tabularnewline
				\toprule[0.75pt] 
		\end{tabular}}
	\end{table*}

	From Table~\ref{tab-ops} we can see that, float addition and float multiplication operations have the highest energy consumption and the longest runtime. More specifically, multiplication operation requires more energy, especially on the Cortex-A15 processor. Low-bit arithmetic has higher efficiency in terms of both energy consumption and latency. However, low-bit multiplication operation still has a relatively high latency of 3 cycles. XNOR is the most efficient operation and takes a single cycle to complete. Compared to other operations, our lookup operation has relatively low energy consumption and requires a latency of only one cycle.

	It can be observed from Table~\ref{tab-network} that our lookup network achieves a $40\%$ energy saving and a $1.6\times$ speedup as compared to the baseline on Cortex-A15. Compared to model 11, our network produces much better performance (92.68/68.97 vs. 91.84/67.60) with over $13\%$ reduction of energy consumption and over $37\%$ reduction of latency on Cortex-A15. Besides, the additional memory consumption of the lookup tables in our LookupNet is very small (72KB). 
	In addition, our lookup operation is also compatible to other network acceleration techniques to achieve further efficiency improvement. {By combining our lookup operation with network quantization technique, our 4-bit lookup network produces higher accuracy than models 8-10 with much lower energy consumption and latency on Cortex-A15.}
	This clearly demonstrates the effectiveness of our lookup operation.
	
\begin{figure}[t]
		\centering
		\includegraphics[width=1\linewidth]{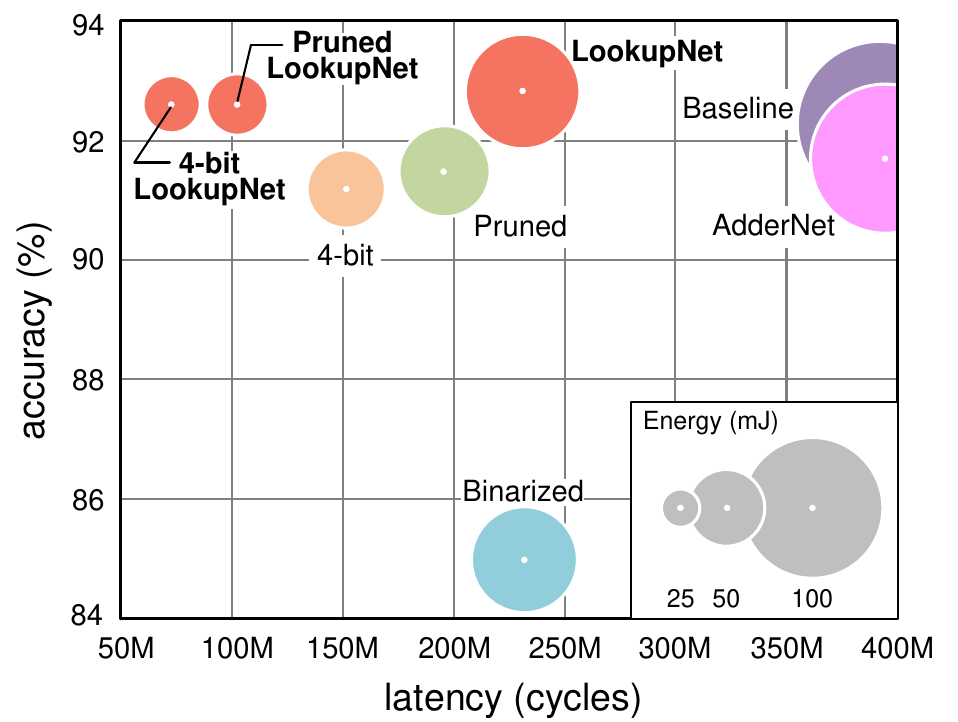}
		\caption{Trade-off between accuracy and efficiency achieved by different methods on Cortex-A15. The size of a circle represents the number of parameters.}
		\label{fig_trade_off}
	\end{figure}

	We further show the trade-off between accuracy and efficiency of different methods in Fig.~\ref{fig_trade_off}. Compared to pruned \cite{Li2020DHP}, quantized \cite{Choi2018Pact}, and binarized models \cite{Zhou2016Dorefa}, our LookupNet produces higher accuracy with competitive efficiency. Compared to AdderNet, our LookupNet achieves higher efficiency in terms of both latency and energy consumption with a notable accuracy improvement. Moreover, network pruning and network quantization techniques can further reduce the computational cost of our LookupNet while maintaining competitive accuracy.

	\noindent \textbf{(2) Practical Results}
	
	{In addition to the aforementioned theoretical analyses, we further conduct experiments to investigate the practical efficiency of our LookupNet on different types of hardwares. \textcolor{black}{Specifically, Nvidia RTX3090, Kirin 810, and Xilinx KV260 are used as the platforms of GPU, mobile CPU processor, and FPGA, respectively. Since network quantization and XNOR operation rely on delicate hardwares (\emph{e.g.}, fixed-point calculator) to achieve speedup, only network pruning methods \cite{Lin2019Towards,Li2020DHP} and AdderNet \cite{Chen2020AdderNet} were included for comparison on these general hardwares. On GPU and mobile CPU, we report memory consumption and inference time of different methods. On FPGA, the numbers of used LUTs and registers with the energy cost are adopted to measure the computational consumption.} Quantitative results are presented in Table~\ref{tab-time}.}
	
	Compared to the baseline, network pruning methods introduce moderate memory reduction and speedup. Since the cost of the model (\emph{e.g.}, model loading and kernel call) is less dominant for larger batch sizes, the efficiency gains are more significant. Meanwhile, AdderNet cannot introduce notable memory reduction or speedup since the cost of float addition operation remains similar to float multiplication operation on these  hardwares (Table~2). In contrast, our LookupNet achieves the highest accuracy with lower memory and computational cost. For example, our lookup operation facilitates VGG-Small to produce a $0.33\times$ memory saving and a $1.78\times$ speedup on GPU for batch size of 512. \textcolor{black}{Moreover, our LookupNet is also compatible with network pruning technique to produce further efficiency gains. Thanks to our FPGA-friendly lookup operation, our LookupNet saves over $50\%$ computational resources as compared to the baseline on FPGA. In addition, network pruning can further facilitate our LookupNet to achieve higher efficiency. Note that, our method produces superior efficiency gains for VGG-Small as compared to ResNet-20 on GPU. This is because, the fragmented structure of ResNet-20 introduces additional overhead on GPU (\emph{e.g.}, kernel launching and synchronization \cite{Ma2018Shufflenet}) that cannot be decreased by our lookup table. Consequently, the advantages of our method cannot be fully exploited on ResNet-20. In contrast, our method achieves significant speedup on VGG-Small. }

	\begin{table*}[t]
		\caption{Top-1 accuracy (\%) achieved on CIFAR-10 and CIFAR-100. ``\#Ops'' represents the number of operations, including addition, multiplication, lookup, etc.
		}
		\label{tab1}
		\centering
		\renewcommand\arraystretch{1.2}
		\setlength{\tabcolsep}{2.3mm}{
			\begin{tabular}{lclccccccccc}
				\toprule[0.75pt] 
				\multirow{2}{*}{Model} & \multicolumn{2}{l}{\multirow{2}{*}{~~~~~~~~~~~~~Method}} & \multirow{2}{*}{\#Ops} & \multicolumn{2}{c}{Energy (mJ)} & \multicolumn{2}{c}{Latency (cycle)} & {\multirow{2}{*}{\tabincell{c}{CIFAR-10\\(\%)}}} & {\multirow{2}{*}{\tabincell{c}{CIFAR-100\\(\%)}}}
				\tabularnewline
				\cmidrule(lr){5-6}\cmidrule(lr){7-8}
				& & & & Cortex-A7 & Cortex-A15 & Cortex-A7 & Cortex-A15
				\tabularnewline
				\toprule[0.75pt]
				\multirow{14}{*}{ResNet-20}
				& & Baseline & 78M & 15.7 & 124.2 & 312M & 390M & 92.25 & 68.14
				\tabularnewline
				\cline{2-10}
				& \multirow{3}{*}{\rotatebox{0}{Pruning}} & DHP \cite{Li2020DHP} & 38M & 7.6 & 60.5 & 152M & 190M & 91.54 & -
				\tabularnewline
				& & FBS \cite{Gao2019Dynamic} & 36M & \textbf{7.2} & 57.3 & 144M & 180M & 90.97 & -
				\tabularnewline
				& & ManiDP \cite{Tang2021Manifold} & 36M & \textbf{7.2} & 57.3 & 144M & 180M & 92.05 & -
				\tabularnewline
				\cline{2-10}
				& \multirow{3}{*}{\rotatebox{0}{Quant.}} & PACT-4bit \cite{Choi2018Pact} & 78M & 8.9 & 49.8 & 156M & 156M& 91.30 & -
				\tabularnewline
				& & QIL-4bit \cite{Jung2019Learning} & 78M & 8.9 & 49.8 & 156M & 156M & 91.52 & -
				\tabularnewline
				& & SLB-4bit \cite{Yang2020Searching} & 78M & 8.9 & 49.8 & 156M & 156M & 91.60 & -
				\tabularnewline
				\cline{2-10}
				& \multirow{4}{*}{\rotatebox{0}{Cheap}} & BNN \cite{Zhou2016Dorefa} & 78M & 10.6 & 72.7 & 195M & 234M & 84.87 & 54.14
				\tabularnewline
				& & DeepShift \cite{Elhoushi2021Deepshift} & 78M & - & - & 195M & 234M & 89.85 & -
				\tabularnewline
				& & ShiftAddNet \cite{You2020Shiftaddnet} & 78M & - & - & 351M & 429M & 85.10 & -
				\tabularnewline
				& & AdderNet \cite{Chen2020AdderNet} & 78M & 15.5 & 114.7 & 312M & 390M & 91.84 & 67.60
				\tabularnewline
				\cline{2-10}
				& \multirow{2}{*}{\rotatebox{0}{Ours}} & LookupNet & 78M & 13.6 & 75.0 & 195M & 234M & \textbf{92.68} & \textbf{68.97}
				\tabularnewline				
				& & LookupNet-4bit & 78M & 9.0 & \textbf{34.5} & \textbf{78M} & \textbf{78M} & {92.51} & {68.35}
				\tabularnewline
				\toprule[0.75pt] 
				\multirow{14}{*}{VGG-Small}
				& & Baseline & 1152M & 231.6 & 1834.6 & 4608M & 5760M & 93.80 & 72.73
				\tabularnewline
				\cline{2-10}
				& \multirow{3}{*}{\rotatebox{0}{Pruning}} & GAL \cite{Lin2019Towards} & 632M & 127.0 & 1006.5 & 2528M & 3160M & 93.42 & - 
				\tabularnewline
				& & HRank \cite{Lin2020HRank} & 400M & 80.4 & 637.0 & 1600M & 2000M & 92.34 & - 
				\tabularnewline
				& & CHIP \cite{Sui2021CHIP} & 384M & \textbf{77.2} & 611.5 & 1536M & 1920M & 93.72 & - 
				\tabularnewline	
				\cline{2-10}		
				& \multirow{3}{*}{\rotatebox{0}{Quant.}} & QIL-4bit \cite{Jung2019Learning} & 632M & 131.3 & 736.1 & 2304M & 2304M & 93.77 & -
				\tabularnewline
				& & SLB-4bit \cite{Yang2020Searching} & 632M & 131.3 & 736.1 & 2304M & 2304M & 93.80 & - 
				\tabularnewline
				& & CPQ-4bit \cite{Lee2021Cluster} & 632M & 131.3 & 736.1 & 2304M & 2304M & 93.23 & -
				\tabularnewline
				\cline{2-10}
				& \multirow{4}{*}{\rotatebox{0}{Cheap}} & BNN \cite{Zhou2016Dorefa} & 632M & 156.1 & 1073.1 & 2880M & 3456M & 89.80 & 65.41
				\tabularnewline
				& & DeepShift \cite{Elhoushi2021Deepshift} & 632M & - & - & 2880M & 3456M & 91.57 & -
				\tabularnewline
				& & ShiftAddNet \cite{You2020Shiftaddnet} & 632M & - & - & 5184M & 6336M & 92.10 & 63.20
				\tabularnewline
				& & AdderNet \cite{Chen2020AdderNet} & 632M & 229.2 & 1694.6 & 4608M & 5760M & 93.72 & 72.64
				\tabularnewline
				\cline{2-10}
				& \multirow{2}{*}{\rotatebox{0}{Ours}} & LookupNet & 632M & 201.0 & 1107.6 & 2880M & 3456M & \textbf{94.14} & \textbf{75.75}
				\tabularnewline
				& & LookupNet-4bit & 632M & 133.6 & \textbf{509.2} & \textbf{1152M} & \textbf{1152M} & 94.10 & 75.24
				\tabularnewline
				\toprule[0.75pt]
		\end{tabular}}
	\end{table*}
	
	\subsection{Experiments on Image Classification}
	
	In this part, we first evaluate our lookup network on the CIFAR-10 and CIFAR-100 datasets. Then, we conduct experiments on the ImageNet dataset.
	
	\subsubsection{Experiments on CIFAR}
	
	\begin{table*}[t]
		\caption{Top-1/Top-5 accuracy achieved on ImageNet for image classification. ``\#Ops'' represents the number of operations, including addition, multiplication, lookup, etc.}
		\label{tab2}
		\centering
		\renewcommand\arraystretch{1.2}
		\setlength{\tabcolsep}{2.8mm}{
			\begin{tabular}{lclccccccccccc}
				\toprule[0.75pt] 
				\multirow{2}{*}{Model} & \multicolumn{2}{l}{\multirow{2}{*}{~~~~~~~~~~~~~Method}} & \multirow{2}{*}{\#Ops}  & \multicolumn{2}{c}{Energy (mJ)} & \multicolumn{2}{c}{Latency (cycle)} & {\multirow{2}{*}{\tabincell{c}{Top-1\\(\%)}}} & {\multirow{2}{*}{\tabincell{c}{Top-5\\(\%)}}}
				\tabularnewline
				\cmidrule(lr){5-6}\cmidrule(lr){7-8}
				&&& & Cortex-A7 & Cortex-A15 & Cortex-A7 & Cortex-A15
				\tabularnewline
				\toprule[0.75pt] 
				\multirow{13}{*}{ResNet-18} 
				& & Baseline & 3356M & 674.6 & 5344.4 & 13424M & 16780M & 69.76 & 89.08
				\tabularnewline
				\cline{2-10}
				& \multirow{3}{*}{\rotatebox{0}{Pruning}} & FBS \cite{Gao2019Dynamic} & 1678M & \textbf{337.3} & 2672.2 & 6712M & 8390M & 68.17 & 88.22
				\tabularnewline
				& & DSA \cite{Ning2020Dsa} & 2014M & 404.8 & 3207.3 & 8056M & 10070M & 68.61 & 88.35
				\tabularnewline
				& & ManiDP \cite{Tang2021Manifold} & 1510M & 303.5 & 2404.7 & 6040M & 7550M & 68.35 & 88.29
				\tabularnewline
				\cline{2-10}
				& \multirow{3}{*}{\rotatebox{0}{Quant.}} & PACT-4bit \cite{Choi2018Pact} & 3356M & 382.6 & 2144.5 & 6712M & 6712M & 69.20 & 89.00
				\tabularnewline
				& & QIL-4bit \cite{Jung2019Learning} & 3356M & 382.6 & 2144.5 & 6712M & 6712M & 68.95 & 88.77
				\tabularnewline
				& & CPQ-4bit \cite{Lee2021Cluster} & 3356M & 382.6 & 2144.5 & 6712M & 6712M & 69.63 & 89.04
				\tabularnewline
				\cline{2-10}
				& \multirow{3}{*}{\rotatebox{0}{Cheap}} & BNN \cite{Zhou2016Dorefa} & 3356M & 454.7 & 3126.1 & 8390M & 10068M & 51.20 & 73.20 
				\tabularnewline
				& & DeepShift \cite{Elhoushi2021Deepshift} & 3356M & - & - & 10068M & 10068M & 69.27 & 89.00  
				\tabularnewline
				& & AdderNet \cite{Chen2020AdderNet} & 3356M & 667.8 & 4936.7 & 13424M & 16780M & 67.00 & 87.60  
				\tabularnewline
				\cline{2-10}
				& \multirow{2}{*}{\rotatebox{0}{Ours}} & LookupNet & 3356M & 585.6 & 3228.5 & 8390M & 10068M & \textbf{70.49} & \textbf{89.70}
				\tabularnewline
				& & {LookupNet-4bit} & 3356M & 389.3 & \textbf{1183.4} & \textbf{3356M} & \textbf{3365M} & 70.25 & 89.52
				\tabularnewline
				\toprule[0.75pt] 
				\multirow{11}{*}{MobileNet-v2} 
				& & Baseline & 536M & 107.7 & 853.6 & 2144M & 2680M & 71.80 & 90.43
				\tabularnewline
				\cline{2-10}		
				& \multirow{3}{*}{\rotatebox{0}{Pruning}} & DMC \cite{Gao2020Discrete} & 354M & 57.9 & 458.6 & 1152M & 1440M & 68.37 & 88.46
				\tabularnewline	
				& & GFP \cite{Liu2021Groupa} & 288M & 54.4 & 431.1 & 1083M & 1353M & 69.16 & 75.74
				\tabularnewline		
				& & ManiDP \cite{Tang2021Manifold} & 261M & \textbf{52.7} & 417.2 & 1048M & 1310M & 69.62 & 89.45
				\tabularnewline
				\cline{2-10}	
				& \multirow{3}{*}{\rotatebox{0}{Quant.}} & PACT-4bit \cite{Choi2018Pact} & 536M & 61.1 & 342.5 & 1072M & 1072M & 61.44 & 82.70
				\tabularnewline	
				& & QIL-4bit \cite{Jung2019Learning} & 536M & 61.1 & 342.5 & 1072M & 1072M & 67.23 & 87.49
				\tabularnewline
				& & CPQ-4bit \cite{Lee2021Cluster} & 536M & 61.1 & 342.5 & 1072M & 1072M & 69.17 & 88.74
				\tabularnewline
				\cline{2-10}
				& \multirow{1}{*}{\rotatebox{0}{Cheap}} & BNN \cite{Phan2020Binarizing} & 536M & 58.4 & 499.8 & 1340M & 1608M & 59.30 & 81.00
				\tabularnewline
				\cline{2-10}
				& \multirow{2}{*}{\rotatebox{0}{Ours}}  & LookupNet & 536M & 93.5 & 515.3 & 1340M & 1608M & {\textbf{70.41}}  & {\textbf{89.53}}
				\tabularnewline
				& & {LookupNet-4bit} & 536M & 62.2 & \textbf{236.9} & \textbf{536M} & \textbf{536M} & 69.75 & 89.30
				\tabularnewline
				\toprule[0.75pt]
		\end{tabular}}
	\end{table*}

	\noindent \textbf{Settings.} 
	The CIFAR-10 and CIFAR-100 datasets \cite{Krizhevsky2009Learning} are two commonly used small-scale datasets for the image classification task.
	The CIFAR-10 dataset contains 50K training images and 10K test images of size $32\times32$ from 10 classes, while the CIFAR-100 dataset comprises of  50K training images and 10K test images of size $32\times32$ from 100 classes. 
	
	We used ResNet-20 \cite{He2016Deep} and VGG-Small \cite{Cai2017Deep} as our baseline networks, and then converted them to lookup networks by replacing convolutional layers with lookup layers. Following \cite{Chen2020AdderNet,Han2020GhostNet}, the first and the last convolutional layers were preserved for fair comparison. 
	During training, the original $32\times32$ images were padded with 4 pixels on each side. Then, $32\times32$ patches were randomly cropped and horizontally flipped. The stochastic gradient descent (SGD) method with momentum of 0.9 was used for optimization. All models were trained for 200 epochs with a mini-batch size of 128 and with the weight decay being set to $5\times10^{-4}$. For both CIFAR-10 and CIFAR-100, the learning rate was initially set to 0.1 and decayed by a factor of 10 at epoch 80 and 160. The gradients were clipped with a maximum L2 norm of 3.

	\noindent \textbf{Performance Evaluation.} We compare our lookup network to three families of methods, including network pruning based methods \cite{Lin2019Towards,Gao2019Dynamic,Li2020DHP,Lin2020HRank,Tang2021Manifold,Sui2021CHIP}, network quantization methods \cite{Choi2018Pact,Jung2019Learning,Yang2020Searching,Lee2021Cluster}, and cheap operation based methods \cite{Zhou2016Dorefa,Chen2020AdderNet,Elhoushi2021Deepshift,You2020Shiftaddnet}. Comparative results are presented in Table~\ref{tab1}. Following \cite{Chen2020AdderNet}, the computational cost in the first and the last layers are omitted when calculating energy consumption and latency since it is significantly less than that in other layers. Besides, the computational cost in BN layers is also omitted since BN layers can be merged into convolutional ones.
	
	For ResNet-20, it can be observed that our LookupNet achieves competitive performance to the baseline on both CIFAR-10 and CIFAR-100. Although network pruning methods \cite{Gao2019Dynamic,Li2020DHP,Tang2021Manifold} have lower energy consumption, these methods suffer inferior performance. Besides, network quantization methods \cite{Choi2018Pact,Jung2019Learning,Yang2020Searching} achieves lower computational complexity using low-bit operations at the cost of an accuracy drop of over $0.6\%$. 
	BNN \cite{Zhou2016Dorefa}, DeepShift \cite{Elhoushi2021Deepshift}, AdderNet \cite{Chen2020AdderNet}, and ShiftAddNet \cite{You2020Shiftaddnet} replace costly multiplication operations with XNOR, shift, and addition operations to achieve higher efficiency. Nevertheless, these methods suffer notable performance loss. In contrast, our LookupNet benefits from the lookup operations to produce much higher accuracy. For example, our LookupNet outperforms AdderNet with significant gains (92.68\%/68.97\% vs. 91.84\%/67.60\%). 
	{In addition, by combining our lookup operation with network quantization technique, our 4-bit lookup network achieves the lowest energy consumption and latency with competitive accuracy (92.51/68.35). This further validates the superiority of our network.}
	For VGG-Small, our LookupNet performs favorably against the baseline while outperforming other methods with significant performance gains. {Moreover, our 4-bit LookupNet also produces the best trade-off between accuracy and efficiency among all approaches.}

	\subsubsection{Experiments on ImageNet}
	
	\noindent\textbf{Settings.}
	The ImageNet (ILSVRC-2012) dataset \cite{Deng2009Imagenet} includes $\sim$1.2M training images and 50K validation images from 1K classes.
	We used ResNet-18 and MobileNet-v2 as the baseline networks to obtain lookup networks. Pre-trained convolutional models were used for initialization. Following \cite{Chen2020AdderNet,Han2020GhostNet}, the first and the last convolutional layers were preserved. 
	
	During training, the original images were resized, cropped to $224\times224$ and randomly flipped horizontally for data augmentation. The SGD method with momentum of 0.9 was used for optimization. The gradients were clipped with a maximum L2 norm of 3. 
	For Resnet-18, its lookup version was trained for 120 epochs with a mini-batch size of 1024. The learning rate was initially set to 0.01 and decayed by a factor of 10 at epoch 30, 60, and 90, with weight decay being set to $1\times10^{-4}$. 
	For MobileNet-v2, {its lookup version was trained for 40 epochs with a mini-batch size of 256. The learning rate was initially set to 0.005 and decayed by a factor of 10 at epoch 10, 20, and 30, with weight decay being set to $4\times10^{-5}$.}
	
	\noindent\textbf{Performance Evaluation.} We compare our lookup network to three families of methods, including network pruning based methods \cite{Gao2019Dynamic,Ning2020Dsa,Gao2020Discrete,Tang2021Manifold,Liu2021Groupa}, network quantization methods \cite{Choi2018Pact,Jung2019Learning,Lee2021Cluster}, and cheap operation based methods \cite{Zhou2016Dorefa,Chen2020AdderNet,Phan2020Binarizing,Elhoushi2021Deepshift}. Comparative results are presented in Table~\ref{tab2}. Following \cite{Chen2020AdderNet}, the computational cost in the first and the last layers are omitted when calculating energy consumption and latency.
	
	It can be observed that our LookupNet produces higher accuracy than other methods. For ResNet-18, we can see that the accuracy of network pruning methods \cite{Gao2019Dynamic,Ning2020Dsa,Tang2021Manifold} and network quantization methods \cite{Choi2018Pact,Jung2019Learning,Lee2021Cluster} is degraded. Using XNOR operation to replace multiplication operation, BNN \cite{Zhou2016Dorefa} suffers limited performance (51.20\%/73.20\%). Although AdderNet improves BNN with notable gains, its accuracy is still inferior to the baseline (67.00\%/87.60\% vs. 69.76\%/89.08\%). In contrast, our LookupNet performs favorably against the baseline and achieves state-of-the-art performance (70.49\%/89.70\%). {With additional network quantization techniques, our lookup network achieves the highest inference efficiency on Cortex-A15 while maintaining state-of-the-art accuracy (70.25\%/89.52\%).} {For MobileNet-v2, our LookupNet produces consistent accuracy improvements against other methods. This further demonstrates that our lookup operation is compatible to lightweight network architectures to achieve higher efficiency.}
	
	\subsection{Experiments on Image Super-Resolution}
	
	In addition to classification task, our lookup network can also be applied to regression tasks, such as image SR. In this section, we conduct experiments to evaluate our lookup network on the image SR task.
	
	\begin{table*}[t]
		\caption{PSNR/SSIM results achieved on four benchmarks for $\times4$ image super-resolution. ``\#Ops'' represents the number of operations, including addition, multiplication, lookup, etc. Results are calculated based on HR images with a resolution of 720p ($1280\times720$).}
		\label{tab4}
		\centering
		\renewcommand\arraystretch{1.2}
		\setlength{\tabcolsep}{0.7mm}{
			\begin{tabular}{lllccccccccc}
				\toprule[0.75pt] 
				\multirow{2}{*}{Model} & \multicolumn{2}{l}{\multirow{2}{*}{~~~~~~~~~~~~~Method}} & \multirow{2}{*}{\#Ops} & \multicolumn{2}{c}{Energy (J)} & \multicolumn{2}{c}{Latency (cycle)} & \multirow{2}{*}{Set5} & \multirow{2}{*}{Set14} & \multirow{2}{*}{B100} & \multirow{2}{*}{Urban100}
				\tabularnewline
				\cmidrule(lr){5-6}\cmidrule(lr){7-8}
				&&&& Cortex-A7 & Cortex-A15 & Cortex-A7 & Cortex-A15
				\tabularnewline
				\toprule[0.75pt] 
				\multirow{10}{*}{EDSR} 
				& & Baseline & 2026G & 407.2 & 3226.4 & 8104G & 10130G & 32.46/0.8968 & 28.80/0.7876 & 27.71/0.7420 & 26.64/0.8033
				\tabularnewline
				\cline{2-12}
				& \multirow{2}{*}{\rotatebox{0}{Pruning}} & Basis \cite{Li2019Learning} & 1306G & 262.5 & 2079.8 & 5224G & 6530G & 31.95/-~~~~~~~~~ & 28.42/-~~~~~~~~~ & 27.46/-~~~~~~~~~ & 25.76/-~~~~~~~~~
				\tabularnewline
				& & DHP \cite{Li2020DHP} & 1246G & 250.8 & 1987.4 & 4992G & 6240G & 31.99/-~~~~~~~~~ & 28.52/-~~~~~~~~~ & 27.53/-~~~~~~~~~ & 25.92/-~~~~~~~~~
				\tabularnewline		
				\cline{2-12}
				& \multirow{2}{*}{\rotatebox{0}{Quant.}} & PACT-4bit \cite{Choi2018Pact} & 2026G & \textbf{231.0} & 1294.6 & 4052G & 4052G & 31.39/0.8834 & 28.10/0.7695 & 27.25/0.7245 & 25.15/0.7535
				\tabularnewline
				& & PAMS-4bit \cite{Li2020PAMS} & 2026G & \textbf{231.0} & 1294.6 & 4052G & 4052G & 31.59/0.8851 & 28.20/0.7725 & 27.32/0.7282 & 25.32/0.7624
				\tabularnewline			
				\cline{2-12}
				& \multirow{2}{*}{\rotatebox{0}{Cheap}} & IBTM \cite{Jiang2021Training} & 2026G & 274.5 & 1887.2 & 5065G & 6078G & 31.84/0.8900 & 28.33/0.7770 & 27.42/0.7320 & 25.54/0.7690
				\tabularnewline
				& & AdderSR \cite{Song2021Addersr} & 2026G & 403.2 & 2980.2 & 8104G & 10130G & 32.13/0.8864 & 28.57/0.7800 & 27.58/0.7368 & 26.33/0.7874
				\tabularnewline
				\cline{2-12}
				& \multirow{2}{*}{\rotatebox{0}{Ours}} & LookupNet & 2026G & 353.5 & 1948.1 & 5065G & 6078G & \textbf{32.41}/\textbf{0.8987} & \textbf{28.78}/\textbf{0.7870} & \textbf{27.71}/\textbf{0.7414} & \textbf{26.59}/\textbf{0.8023}
				\tabularnewline
				
				& & LookupNet-4bit & 2026G & 235.0 & \textbf{895.5} & \textbf{2026G} & \textbf{2026G} & 32.38/0.8986 & 28.75/0.7869 & 27.71/0.7412 & 26.56/0.8019
				\tabularnewline
				\toprule[0.75pt] 
				\multirow{6}{*}{VDSR} 
				& & Baseline & 1140G & 229.1 & 1815.5 & 4560G & 5700G & 31.35/0.8838 & 28.01/0.7674 & 27.29/0.7251 & 25.18/0.7524
				\tabularnewline
				\cline{2-12}				
				& \multirow{2}{*}{\rotatebox{0}{Cheap}} & IBTM \cite{Jiang2021Training} & 1140G & 154.5 & 1061.9 & 2850G & 3420G & 31.06/0.8770 & 27.85/0.7620 & 27.07/0.7180 & 24.88/0.7400
				\tabularnewline
				& & AdderSR \cite{Song2021Addersr} & 1140G & 226.9 & 1676.9 & 4560G & 5700G & 31.27/0.8762 & 27.93/0.7630 & 27.25/0.7229 & 25.09/0.7445
				\tabularnewline
				\cline{2-12}
				& \multirow{2}{*}{\rotatebox{0}{Ours}} & LookupNet & 1140G & 198.9 & 1096.1 & 2850G & 3420G & \textbf{31.59}/\textbf{0.8883} & \textbf{28.22}/\textbf{0.7736} & \textbf{27.34}/\textbf{0.7282} & \textbf{25.46}/\textbf{0.7652}
				\tabularnewline				
				& & LookupNet-4bit & 1140G & \textbf{132.2} & \textbf{503.9} & \textbf{1140G} & \textbf{1140G} & 31.49/0.8864 & 28.20/0.7724 & 27.32/0.7263 & 25.43/0.7625
				\tabularnewline
				\toprule[0.75pt]
		\end{tabular}}
	\end{table*}
	
	\begin{figure*}[t]
		\centering
		\includegraphics[width=1\linewidth]{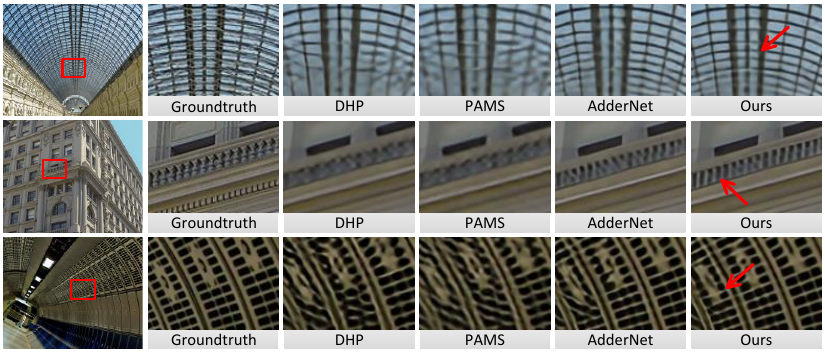}
		\caption{Visual results produced by different methods for $\times4$ image super-resolution. EDSR is used as the baseline model. Note that, since the official code for AdderNet \cite{Song2021Addersr} is not released, we use our implementation to produce its SR results.}
		\label{fig_SR}
	\end{figure*}
	
	\noindent\textbf{Settings.}
	We used 800 training images in DIV2K \cite{Agustsson2017NTIRE} as the training set and included four benchmark datasets (Set5 \cite{Bevilacqua2012Low}, Set14 \cite{Zeyde2010Single}, B100 \cite{Martin2001database}, and Urban100 \cite{Huang2015Single}) for evaluation.
	EDSR \cite{Lim2017Enhanced} and VDSR \cite{Kim2016Accurate} were used as baselines to obtain our lookup networks. Pre-trained convolutional models were used for initialization\footnote{For EDSR, official pre-trained model was employed. For VDSR, we trained a convolutional model for initialization using our implementation.}. Following \cite{Song2021Addersr}, only convolutional layers in backbone blocks were replaced with lookup layers, with the first and the last convolutional layers being preserved. 
	
	During training, 12 low-resolution patches of size $48\times48$ and their high-resolution counterparts were randomly cropped. Then, data augmentation was performed through random rotation and random flipping. The Adam \cite{Kingma2015Adam} method with $\beta_1=0.9$ and $\beta_2=0.99$ was used for optimization. The $L_1$ loss between SR results and HR images was used as the loss function. The initial learning rate was set to $1\times10^{-4}$ and halved every 10 epochs. All models were trained for 40 epochs.
	
	\begin{table*}[t]
		\caption{Overall accuracy achieved on ModelNet40 for point cloud classification. ``\#Ops'' represents the number of operations, including addition, multiplication, lookup, etc. Results are calculated using 1K points as input.}
		\label{tab_point}
		\centering
		\renewcommand\arraystretch{1.2}
		\setlength{\tabcolsep}{4mm}{
			\begin{tabular}{lllccccccccccc}
				\toprule[0.75pt] 
				\multirow{2}{*}{Model} & \multicolumn{2}{l}{\multirow{2}{*}{~~~~~~~~~~~~~Method}} & \multirow{2}{*}{\#Ops} & \multicolumn{2}{c}{Energy (mJ)} & \multicolumn{2}{c}{Latency (cycle)} & {\multirow{2}{*}{\tabincell{c}{Acc\\(\%)}}}
				\tabularnewline
				\cmidrule(lr){5-6}\cmidrule(lr){7-8}
				& & & & Cortex-A7 & Cortex-A15 & Cortex-A7 & Cortex-A15
				\tabularnewline
				\toprule[0.75pt] 
				\multirow{8}{*}{PointNet} 
				& & Baseline & 820M & 164.8 & 1305.9 & 3280M & 4100M & 90.8
				\tabularnewline
				\cline{2-9}
				& \multirow{2}{*}{\rotatebox{0}{Quant.}} & PACT-4bit \cite{Choi2018Pact} & 820M & \textbf{93.5} & 524.0 & 1640M & 1640M & 89.4
				\tabularnewline
				& & QIL-4bit \cite{Jung2019Learning} & 820M & \textbf{93.5} & 524.0 & 1640M & 1640M & 89.7
				\tabularnewline
				\cline{2-9}
				& \multirow{2}{*}{\rotatebox{0}{Cheap}} & BNN \cite{Qin2021BiPointNet} & 820M & 111.1 & 763.8 & 2050M & 2460M & 85.6
				\tabularnewline
				& & AdderNet \cite{Chen2020AdderNet} & 820M  & 163.2 & 1206.2 & 3280M & 4100M & 89.2
				\tabularnewline
				\cline{2-9}
				& \multirow{2}{*}{\rotatebox{0}{Ours}} & LookupNet & 820M  & 143.1 & 788.4 & 2050M & 2460M & \textbf{90.5}
				\tabularnewline			
				& & LookpNet-4bit & 820M  & 95.1 & \textbf{362.4} & \textbf{820M} & \textbf{820M} & 89.9
				\tabularnewline
				\toprule[0.75pt] 
				\multirow{8}{*}{PointNet++} 
				& & Baseline & 1422M & 285.8 & 2264.5 & 5688M & 7110M & 92.8
				\tabularnewline
				\cline{2-9}		
				& \multirow{2}{*}{\rotatebox{0}{Quant.}} & PACT-4bit \cite{Choi2018Pact} & 1422M & \textbf{162.1} & 908.7 & 2844M & 2844M & 92.3
				\tabularnewline
				& & QIL-4bit \cite{Jung2019Learning} & 1422M & \textbf{162.1} & 908.7 & 2844M & 2844M & 92.6
				\tabularnewline
				\cline{2-9}		
				& \multirow{2}{*}{\rotatebox{0}{Cheap}} & BNN \cite{Qin2021BiPointNet} & 1422M & 192.7 & 1324.6 & 3555M & 4266M & 87.8
				\tabularnewline
				& & AdderNet \cite{Chen2020AdderNet} & 1422M & 283.0 & 2091.8 & 5688M & 7110M & 90.8
				\tabularnewline
				\cline{2-9}
				& \multirow{2}{*}{\rotatebox{0}{Ours}} & LookupNet & 1422M  & 248.1 & 1367.3 & 3555M & 4266M & \textbf{92.7}
				\tabularnewline				
				& & LookpNet-4bit & 1422M  & 164.9 & \textbf{628.5} & \textbf{1422M} & \textbf{1422M} & 92.2
				\tabularnewline
				\toprule[0.75pt] 
		\end{tabular}}
	\end{table*}
	
	\noindent\textbf{Performance Evaluation.}
	We compare our lookup network to three families of methods, including network pruning based methods \cite{Li2019Learning,Li2020DHP}, network quantization methods \cite{Choi2018Pact,Li2020PAMS}, and cheap operation based methods \cite{Jiang2021Training,Song2021Addersr}. Quantitative results are presented in Table~\ref{tab4} while qualitative results are illustrated in Fig.~\ref{fig_SR}. Following \cite{Chen2020AdderNet}, the computational cost in the first and the last layers are omitted when calculating energy consumption and latency. 
	
	For EDSR, we can see that our LookupNet produces comparable performance to the baseline and significantly outperforms other methods on different datasets in terms of PSNR and SSIM. By pruning redundant parameters in EDSR, network pruning methods \cite{Li2019Learning,Li2020DHP} reduce the number of operations at the cost of notable performance drop. Although network quantization methods \cite{Choi2018Pact,Li2020PAMS} reduce the computational complexity of EDSR by using low-bit operations, these methods suffer relatively low performance. Using cheap operations to replace costly multiplication operation, IBTM \cite{Jiang2021Training} and AdderSR \cite{Song2021Addersr} produce promising efficiency with competitive performance. Benefited from the lookup operations, our LookupNet achieves significant performance gains as compared to these two methods. For example, our LookupNet produces a PSNR improvement of 0.26dB against AdderSR on Urban100. 
	{By quantizing the LookupNet, our network further achieves over $2\times$ reduction in terms of both energy consumption (895.5J vs. 2131.4J) and latency (2026G vs. 6078G) while maintaining comparable performance (26.56 vs. 26.59 on Urban100).}
	For VDSR, our LookupNet achieves competitive performance to the baseline and produces the highest PSNR/SSIM scores among all methods. {Besides, our 4-bit LookupNet achieves a better trade-off between accuracy and efficiency.}
	
	From Fig.~\ref{fig_SR} we can further see that our LookupNet produces results with clearer and finer details, such as the rails in the second row and the grids in the third row. The better perceptual quality of our results further demonstrates the effectiveness of our LookupNet. 
	
	\subsection{Experiments on Point Cloud Classification}
	
	Apart from image processing tasks, our lookup network can also be extended to other modalities such as irregular point clouds. In this section, evaluation experiments are conducted on the point cloud classification task.
	
	\noindent\textbf{Settings.}
	We used the ModelNet40 dataset \cite{Wu20153d} to evaluate our method on point cloud classification. This dataset contains 12311 meshed CAD models from 40 categories. 
	PointNet \cite{Qi2017Pointnet} and PointNet++ \cite{Qi2017PointNet++} were adopted as baselines to obtain our lookup networks. For fair comparison with \cite{Qin2021BiPointNet}, the first and the last convolutional layers were preserved. 
	Following \cite{Qi2017PointNet++,Xu2021PAConv}, we sampled 1024 points from each object as the input of the network. 
	During training, batch size was set to 24. The Adam \cite{Kingma2015Adam} method with $\beta_1=0.9$ and $\beta_2=0.99$ was used for optimization. The cross-entropy loss was used as the loss function. The learning rate was initialized as 0.001 and multiplied with 0.7 after every 20 epochs. All models were trained for 200 epochs. 
	
	\noindent\textbf{Performance Evaluation.} 
	We compare our lookup network to two families of methods, including network quantization methods \cite{Choi2018Pact,Jung2019Learning} and cheap operation based methods \cite{Chen2020AdderNet,Qin2021BiPointNet}. Table~\ref{tab_point} presents the results achieved by different methods. Following \cite{Chen2020AdderNet}, the computational cost in the first and the last layers are omitted when calculating energy consumption and latency. 
	
	As we can see, our LookupNet achieves the best performance for both PointNet and PointNet++. For PointNet, network quantization methods \cite{Choi2018Pact,Jung2019Learning} suffer over $1\%$ accuracy loss when representing float values with 4-bit values. Using XNOR and addition operations to replace multiplications, BNN \cite{Qin2021BiPointNet} and AdderNet \cite{Chen2020AdderNet} also produce severe performance drop. In contrast, our LookupNet achieves competitive accuracy to the baseline (90.5 vs. 90.8) and significantly surpasses other methods. {Meanwhile, our 4-bit LookupNet produces very competitive results with much lower cost.}
	For PointNet++, our LookupNet also performs favorably against the baseline and produces higher accuracy than other methods. For example, our LookupNet achieves an accuracy improvement of $2.1\%$ as compared to AdderNet. This clearly demonstrates the effectiveness of our lookup operation.

	\section{Conclusion}
	\label{Sec5}
	
	In this paper, we introduce a simple yet efficient lookup operation for neural networks. Our lookup operation uses activation and weight values as indices to find their corresponding values in a 2D lookup table rather than calculate their multiplication. Our lookup operation is differentiable and well compatible to different types of operations. We construct lookup networks using addition and lookup operations for image classification, image SR, and point cloud classification tasks. Extensive experiments show that our lookup networks benefit from the lookup operations to achieve state-of-the-art performance in terms of both accuracy and efficiency.
	
	\section{Acknowledgments}
	{This work is partially supported by the Jilin Province Science and Technology Development Projects (No. 20250102209JC), the National Natural Science Foundation of China (No. 62301601, U20A20185, 62372491), the Guangdong Basic and Applied Basic Research Foundation (2022B1515020103, 2023B1515120087), the Science and Technology Research Projects of the Education Office of Jilin Province (No. JJKH20251951KJ), the Special Financial Grant from China Postdoctoral Science	Foundation (No. 2025T180433), and the Science and Technology Planning Project of Key Laboratory of Advanced IntelliSense Technology, Guangdong Science and Technology Department (No. 2023B1212060024).
	}
	
	
	\bibliographystyle{unsrt}
	\bibliographystyle{IEEEtran}

	\ifCLASSOPTIONcaptionsoff
	
	\newpage
	\fi
	
	
	
	

	\begin{IEEEbiography}[{\includegraphics[width=1in,height=1.25in,clip]{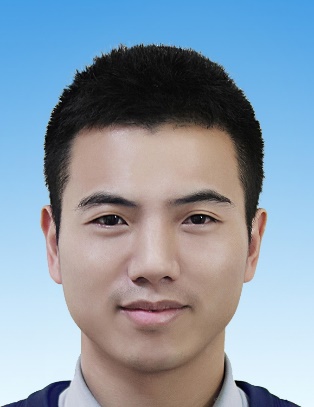}}]
		{Yulan Guo} is a full Professor with the School of Electronics and Communication Engineering, Sun Yat-sen University. His research interests lie in spatial intelligence and 3D vision, particularly in 3D reconstruction, point cloud understanding, and robot systems. He has authored over 200 articles at highly referred journals and conferences. He served as a Senior Area Editor for IEEE Transactions on Image Processing, and an Associate Editor for the Visual Computer, and Computers \& Graphics. He also served as an area chair for CVPR 2025/2023/2021, ICCV 2025/2021, ECCV 2024, NeurIPS 2024, and ACM Multimedia 2021. He organized over 10 workshops, challenges, and tutorials in prestigious conferences such as CVPR, ICCV, ECCV, and 3DV. He is a Senior Member of IEEE and ACM.
	\end{IEEEbiography}
	\vspace{-1.1cm}
	\begin{IEEEbiography}[{\includegraphics[width=1in,height=1.25in,clip]{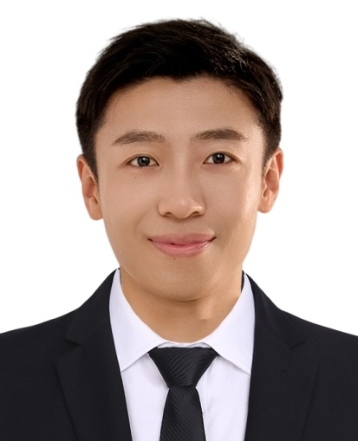}}]
		{Longguang Wang} received the B.E. degree in Electrical Engineering from Shandong University (SDU), Jinan, China, in 2015, and the Ph.D. degree in Information and Communication Engineering from National University of Defense Technology (NUDT), Changsha, China, in 2022. His current research interests include low-level vision and 3D vision.
	\end{IEEEbiography}
	\vspace{-1.1cm}
		\begin{IEEEbiography}[{\includegraphics[width=1in,height=1.25in,clip]{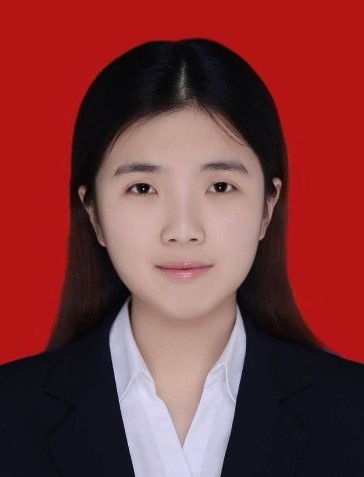}}]
		{Wendong Mao} received the B.S. degree in information engineering from Jilin University, Changchun, China, in 2018, and the Ph.D. degree in information and communication engineering from Nanjing University, China, in 2023. She is currently an assistant professor at the College of Integrated Circuits of Sun Yat-sen University, Shenzhen, China. She was a Visiting Student with the Wangxuan Institute of Computer Technology, Peking University, Beijing, in 2019. Her current research interests include image/video processing algorithm, 3D vision and very large scale integration (VLSI) design for deep learning. She received the IEEE ISVLSI 2022 Best Paper Award. She also serves as the reviewer of various journals and conferences, including TCAS-I, TCAS-II, TNNLS, TVLSI, ISCAS, etc.
	\end{IEEEbiography}
	\vspace{-1.1cm}
	\begin{IEEEbiography}[{\includegraphics[width=1in,height=1.25in,clip]{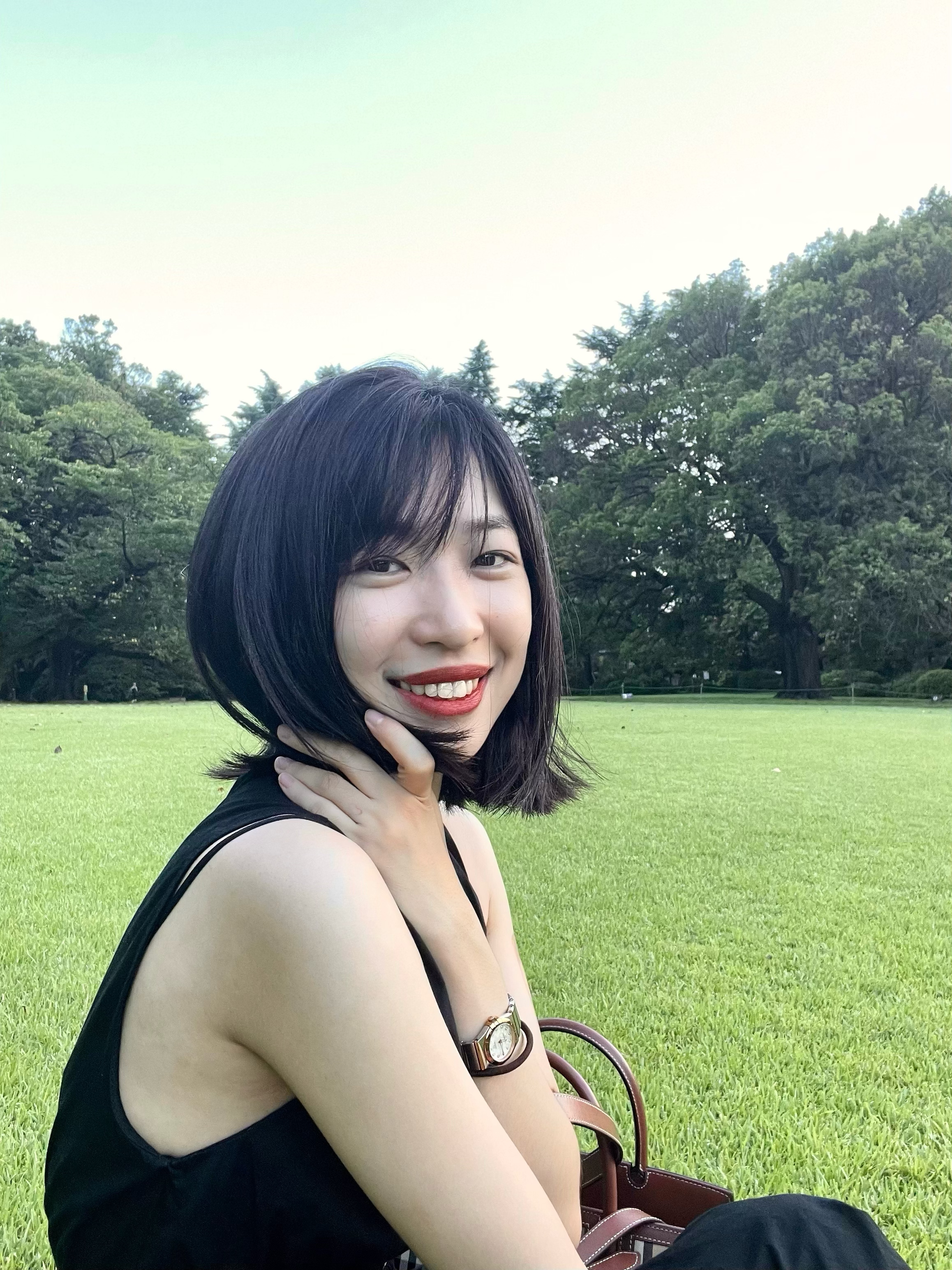}}]
		{Xiaoyu Dong} is a Postdoctoral Researcher at The University of Tokyo. She obtained her Ph.D. degree from The University of Tokyo in 2024 and her M.Eng. degree from Harbin Engineering University in 2021. From April 2021 to September 2024, she worked as a Junior Research Associate at RIKEN AIP. She is a recipient of the 2022 RIKEN Ohbu Award (Researcher Incentive Award). Her research interests include Computational Imaging, Multi-Modal Vision, and 3D Vision.
	\end{IEEEbiography}
	\vspace{-1.1cm}
	\begin{IEEEbiography}[{\includegraphics[width=1in,height=1.25in,clip,keepaspectratio]{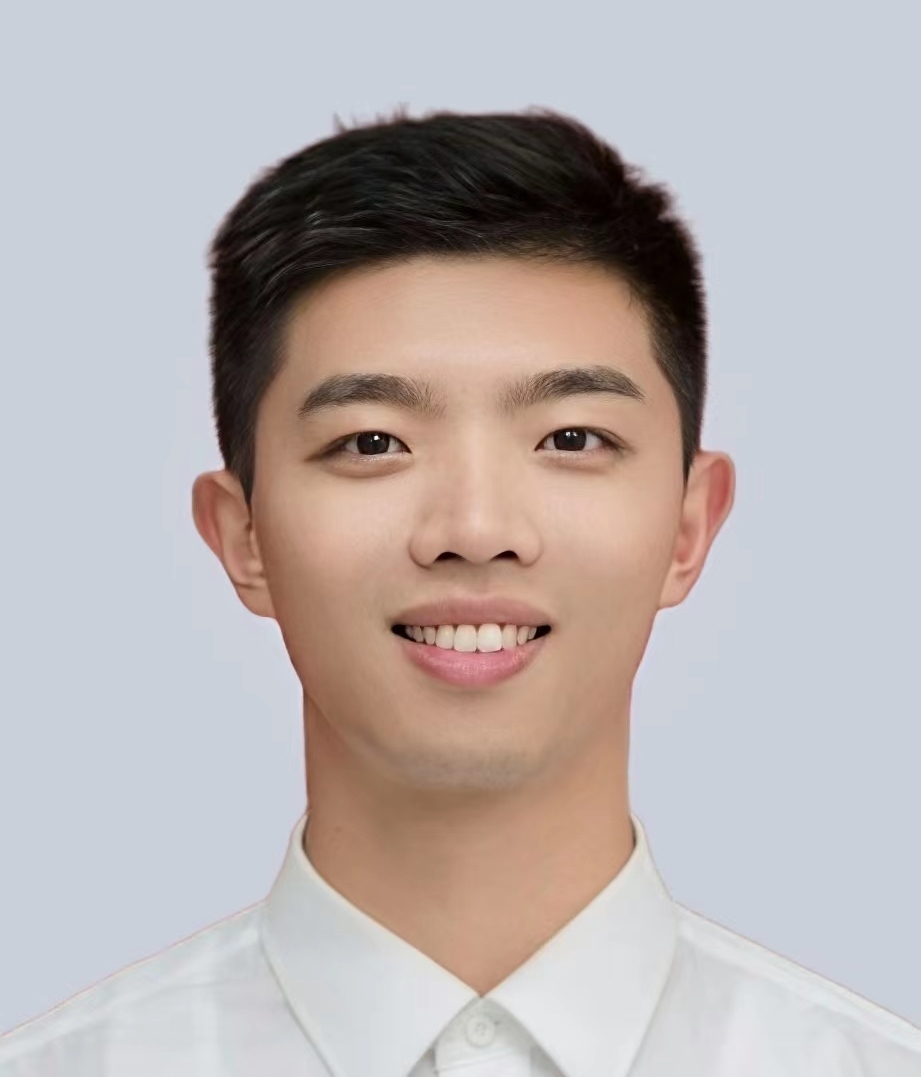}}]
		{Yingqian Wang} received the B.E. degree in electrical engineering from Shandong University, Jinan, China, in 2016, and the M.E. and Ph.D. degrees in information and communication engineering from the National University of Defense Technology (NUDT), Changsha, China, in 2018 and 2023, respectively. He is currently an Assistant Professor and Graduate Supervisor with the College of Electronic Science and Technology, NUDT. His research interests center on computational photography and low-level vision, with a particular focus on light field image processing and image super-resolution.
	\end{IEEEbiography}
	\vspace{-1.1cm}
	\begin{IEEEbiography}[{\includegraphics[width=1in,height=1.25in,clip]{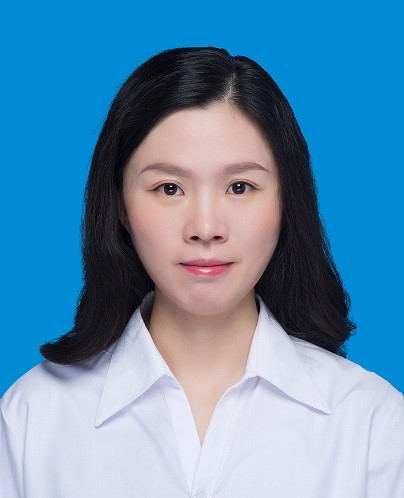}}]
		{Li Liu} received the Ph.D. degree in information and communication engineering from the National University of Defense Technology (NUDT), China, in 2012. During her Ph.D. study, she spent more than two years as a Visiting Student at the University of Waterloo, Canada, from 2008 to 2010. From 2015 to 2016, she spent ten months visiting the Multimedia Laboratory at the Chinese University of Hong Kong. From 2016.12 to 2018.11, she worked as a senior researcher at the Machine Vision Group at the University of Oulu, Finland. Her current research interests include Computer Vision, Machine Learning, Artificial Intelligence, Trustworthy AI, Synthetic Aperture Radar. Her papers have currently over 7500+ citations in Google Scholar. 
	\end{IEEEbiography}
	\vspace{-1.1cm}
	\begin{IEEEbiography}[{\includegraphics[width=1in,height=1.25in,clip]{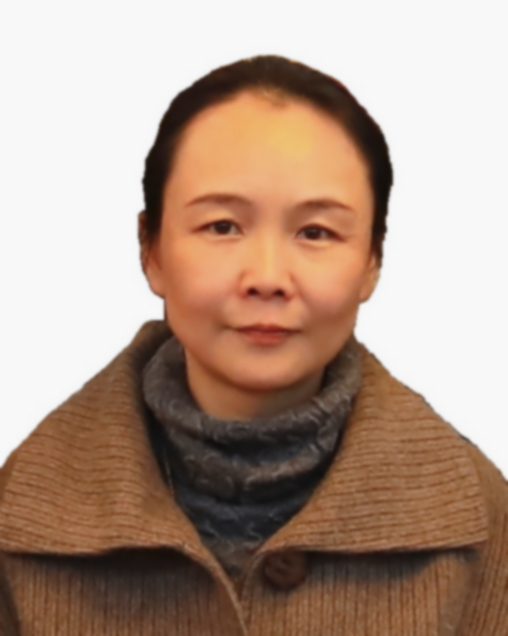}}]
		{Wei An} received the Ph.D. degree from the National University of Defense Technology (NUDT), Changsha, China, in 1999. She was a Senior Visiting Scholar with the University of Southampton, Southampton, U.K., in 2016. She is currently a Professor with the College of Electronic Science and Technology, NUDT. She has authored or co-authored over 100 journal and conference publications. Her current research interests include signal processing and image processing.
	\end{IEEEbiography}
\end{document}